\documentclass[letterpaper, 10 pt, conference]{ieeeconf}  % Comment this line out if you need a4paper

\IEEEoverridecommandlockouts                              % This command is only needed if 
\usepackage{amsmath} % assumes amsmath package installed
\usepackage{amssymb}  % assumes amsmath package installed
\usepackage{graphicx}
\usepackage{hyphenat}
\usepackage[table]{xcolor}
\usepackage[font=small,belowskip=0pt, aboveskip=6pt]{caption}
\usepackage{subcaption}

\usepackage{multirow}
\usepackage{booktabs,amsfonts,dcolumn}
\usepackage{tabularx}
\usepackage{makecell}
\usepackage{hyperref}
\usepackage{comment}
\usepackage{diagbox}
\usepackage{multirow}
\usepackage{calrsfs}

\usepackage[style=ieee,mincitenames=1,maxcitenames=2,maxbibnames=10,doi=false,isbn=false,url=false,eprint=false]{biblatex}

\usepackage{xspace}
\usepackage[group-separator={,}]{siunitx}

\DeclareMathAlphabet{\pazocal}{OMS}{zplm}{m}{n}

\long\def\invis#1{}

\newcommand\eq[1]{Eq. #1}
\newcommand\fig[1]{Fig. #1}
\newcommand\sect[1]{Section #1}

\makeatletter
\DeclareRobustCommand\onedot{\futurelet\@let@token\@onedot}
\def\@onedot{\ifx\@let@token.\else.\null\fi\xspace}

\def\etc{\emph{etc}\onedot} 
 
\def\etal{\emph{et al}\onedot}
\makeatother

\title{\LARGE \bf
Enhancing Visual Inertial SLAM with Magnetic Measurements
}

\author{Bharat Joshi and Ioannis Rekleitis% <-this % stops a space
\thanks{\noindent\rule{5cm}{0.5pt}}
\thanks{The authors are with the Department of Computer Science and Engineering, University of South Carolina, Columbia, SC 29208, USA. {\tt\small bjoshi@email.sc.edu,  yiannisr@cse.sc.edu.}} %
\thanks{This research has been supported in part by the National Science Foundation under grants 1943205 and  2024741. The authors also acknowledge the help of the Woodville Karst Plain Project (WKPP), El Centro Investigador del Sistema Acuífero de Quintana Roo A.C. (CINDAQ), Ricardo Constantino, Project Baseline, and Evan Kornacki in collecting data, providing access to challenging underwater caves, and mentoring us in underwater cave exploration. The authors are also grateful for equipment support by Halcyon Dive Systems, Teledyne FLIR LLC, and KELDAN GmbH lights.}% <-this % stops a space
}

\begin{document}

% $ biblatex auxiliary file $
% $ biblatex bbl format version 3.2 $
% Do not modify the above lines!
%
% This is an auxiliary file used by the 'biblatex' package.
% This file may safely be deleted. It will be recreated by
% biber as required.
%
\begingroup
\makeatletter
\@ifundefined{ver@biblatex.sty}
  {\@latex@error
     {Missing 'biblatex' package}
     {The bibliography requires the 'biblatex' package.}
      \aftergroup }
  {}
\endgroup

\refsection{0}
  \datalist[entry]{none/global//global/global}
    \entry{RahmanOceans2018}{inproceedings}{}
      \name{author}{4}{}{%
        {{hash=99f26b0ff5ddff59d005db7acd08100b}{%
           family={Rahman},
           familyi={R\bibinitperiod},
           given={Sharmin},
           giveni={S\bibinitperiod}}}%
        {{hash=c466b011cd605f3f1c67570732be8b67}{%
           family={Karapetyan},
           familyi={K\bibinitperiod},
           given={Nare},
           giveni={N\bibinitperiod}}}%
        {{hash=ca5720cb4fb280292a8f661739855460}{%
           family={{Quattrini Li}},
           familyi={Q\bibinitperiod},
           given={Alberto},
           giveni={A\bibinitperiod}}}%
        {{hash=b47c2dc8dc3c61fadae36fa40955b725}{%
           family={Rekleitis},
           familyi={R\bibinitperiod},
           given={Ioannis},
           giveni={I\bibinitperiod}}}%
      }
      \list{organization}{1}{%
        {IEEE}%
      }
      \strng{namehash}{19daca79b3da14b70f0ca0897926824c}
      \strng{fullhash}{161fa5b9b72f5710c9cbf52d41747d12}
      \strng{bibnamehash}{161fa5b9b72f5710c9cbf52d41747d12}
      \strng{authorbibnamehash}{161fa5b9b72f5710c9cbf52d41747d12}
      \strng{authornamehash}{19daca79b3da14b70f0ca0897926824c}
      \strng{authorfullhash}{161fa5b9b72f5710c9cbf52d41747d12}
      \field{extraname}{1}
      \field{sortinit}{1}
      \field{sortinithash}{4f6aaa89bab872aa0999fec09ff8e98a}
      \field{labelnamesource}{author}
      \field{labeltitlesource}{title}
      \field{abstract}{This paper presents the design, development, and application of a sensor suite, made with the explicit purpose of localizing and mapping in underwater environments. The design objectives of such an underwater sensor rig include simplicity of carrying, ease of operation in different modes, and data collection. The rig is equipped with stereo camera, inertial measurement unit (IMU), mechanical scanning sonar, and depth sensor. The electronics are enclosed in a water-proof PVC tube tested to sixty meters. The contribution of this paper is twofold: first, we open-source the design providing detailed instructions that are made available online; second, we discuss lessons learned as well as some successful applications where the presented sensor suite has been operated by divers.}
      \field{booktitle}{MTS/IEEE OCEANS - Charleston}
      \field{label}{N19}
      \field{title}{A Modular Sensor Suite for Underwater Reconstruction}
      \field{year}{2018}
      \field{pages}{1\bibrangedash 6}
      \range{pages}{6}
      \verb{file}
      \verb RahmanOceans2018.pdf
      \endverb
    \endentry
    \entry{okvis}{article}{}
      \name{author}{5}{}{%
        {{hash=25564f7f36f46be27f7326210cc5c2ad}{%
           family={Leutenegger},
           familyi={L\bibinitperiod},
           given={Stefan},
           giveni={S\bibinitperiod}}}%
        {{hash=f1f30ae8b7d1975ffa2e3f2c80a78a42}{%
           family={Lynen},
           familyi={L\bibinitperiod},
           given={Simon},
           giveni={S\bibinitperiod}}}%
        {{hash=ee31dafdad04c8baf1a0304f3e9cb39a}{%
           family={Bosse},
           familyi={B\bibinitperiod},
           given={Michael},
           giveni={M\bibinitperiod}}}%
        {{hash=6ee8958bbf32527489012d8ca7c95ee3}{%
           family={Siegwart},
           familyi={S\bibinitperiod},
           given={Roland},
           giveni={R\bibinitperiod}}}%
        {{hash=9f26483b21e967b59019c49805811483}{%
           family={Furgale},
           familyi={F\bibinitperiod},
           given={Paul},
           giveni={P\bibinitperiod}}}%
      }
      \strng{namehash}{143a0baf4e18f8ed752f9e514de1fd5d}
      \strng{fullhash}{61f7f94783f280680ee249d487740425}
      \strng{bibnamehash}{61f7f94783f280680ee249d487740425}
      \strng{authorbibnamehash}{61f7f94783f280680ee249d487740425}
      \strng{authornamehash}{143a0baf4e18f8ed752f9e514de1fd5d}
      \strng{authorfullhash}{61f7f94783f280680ee249d487740425}
      \field{sortinit}{2}
      \field{sortinithash}{8b555b3791beccb63322c22f3320aa9a}
      \field{labelnamesource}{author}
      \field{labeltitlesource}{title}
      \field{journaltitle}{The International Journal of Robotics Research}
      \field{number}{3}
      \field{title}{Keyframe-based visual–inertial odometry using nonlinear optimization}
      \field{volume}{34}
      \field{year}{2015}
      \field{pages}{314\bibrangedash 334}
      \range{pages}{21}
      \verb{doi}
      \verb 10.1177/0278364914554813
      \endverb
    \endentry
    \entry{qin2017vins_mono}{article}{}
      \name{author}{3}{}{%
        {{hash=c6f16bc781b2b3952102ac95e566848a}{%
           family={Qin},
           familyi={Q\bibinitperiod},
           given={Tong},
           giveni={T\bibinitperiod}}}%
        {{hash=de0189069ec807597ed74f1c50f193a1}{%
           family={Li},
           familyi={L\bibinitperiod},
           given={Peiliang},
           giveni={P\bibinitperiod}}}%
        {{hash=0d6f0106ed13aa88a0ff49e841f0fead}{%
           family={Shen},
           familyi={S\bibinitperiod},
           given={Shaojie},
           giveni={S\bibinitperiod}}}%
      }
      \strng{namehash}{8e0232b6f28e3805eab5bc1c1db20c74}
      \strng{fullhash}{ade44334fdbd0306897ff8a730c2acba}
      \strng{bibnamehash}{ade44334fdbd0306897ff8a730c2acba}
      \strng{authorbibnamehash}{ade44334fdbd0306897ff8a730c2acba}
      \strng{authornamehash}{8e0232b6f28e3805eab5bc1c1db20c74}
      \strng{authorfullhash}{ade44334fdbd0306897ff8a730c2acba}
      \field{sortinit}{3}
      \field{sortinithash}{ad6fe7482ffbd7b9f99c9e8b5dccd3d7}
      \field{labelnamesource}{author}
      \field{labeltitlesource}{title}
      \field{journaltitle}{IEEE Transactions on Robotics}
      \field{number}{4}
      \field{title}{VINS-Mono: A Robust and Versatile Monocular Visual-Inertial State Estimator}
      \field{volume}{34}
      \field{year}{2018}
      \field{pages}{1004\bibrangedash 1020}
      \range{pages}{17}
    \endentry
    \entry{forster2016manifold}{article}{}
      \name{author}{4}{}{%
        {{hash=a6b6e778fcd0ba37d9fef8a433471203}{%
           family={Forster},
           familyi={F\bibinitperiod},
           given={Christian},
           giveni={C\bibinitperiod}}}%
        {{hash=09ed5bfe663f495a0cb6828e722b7919}{%
           family={Carlone},
           familyi={C\bibinitperiod},
           given={Luca},
           giveni={L\bibinitperiod}}}%
        {{hash=44074d3cb57c174a8b15ccc9a576ac0d}{%
           family={Dellaert},
           familyi={D\bibinitperiod},
           given={Frank},
           giveni={F\bibinitperiod}}}%
        {{hash=dac21ab4ede215439fcc6b051be53a11}{%
           family={Scaramuzza},
           familyi={S\bibinitperiod},
           given={Davide},
           giveni={D\bibinitperiod}}}%
      }
      \list{publisher}{1}{%
        {IEEE}%
      }
      \strng{namehash}{c160fa34da31c9b364cb29944b477266}
      \strng{fullhash}{df880edafe55d50cae3cadb735b5bd7e}
      \strng{bibnamehash}{df880edafe55d50cae3cadb735b5bd7e}
      \strng{authorbibnamehash}{df880edafe55d50cae3cadb735b5bd7e}
      \strng{authornamehash}{c160fa34da31c9b364cb29944b477266}
      \strng{authorfullhash}{df880edafe55d50cae3cadb735b5bd7e}
      \field{sortinit}{3}
      \field{sortinithash}{ad6fe7482ffbd7b9f99c9e8b5dccd3d7}
      \field{labelnamesource}{author}
      \field{labeltitlesource}{title}
      \field{journaltitle}{IEEE Transactions on Robotics}
      \field{number}{1}
      \field{title}{On-manifold preintegration for real-time visual--inertial odometry}
      \field{volume}{33}
      \field{year}{2016}
      \field{pages}{1\bibrangedash 21}
      \range{pages}{21}
    \endentry
    \entry{Florida2010}{report}{}
      \list{institution}{2}{%
        {Florida Ocean}%
        {Coastal Council}%
      }
      \list{location}{1}{%
        {Tallahasee, FL}%
      }
      \field{sortinit}{4}
      \field{sortinithash}{9381316451d1b9788675a07e972a12a7}
      \field{labeltitlesource}{title}
      \field{title}{{Climate Change and Sea-Level Rise in Florida}}
      \field{type}{techreport}
      \field{year}{2010}
    \endentry
    \entry{ZexuanReports2016}{article}{}
      \name{author}{4}{}{%
        {{hash=dffdb3625b1ca807c59a560ab5c46766}{%
           family={Xu},
           familyi={X\bibinitperiod},
           given={Zexuan},
           giveni={Z\bibinitperiod}}}%
        {{hash=1b98df9e7b30ad7917707150ccd6f8fd}{%
           family={Bassett},
           familyi={B\bibinitperiod},
           given={Seth\bibnamedelima Willis},
           giveni={S\bibinitperiod\bibinitdelim W\bibinitperiod}}}%
        {{hash=a7e218b0f81ee2117192c9cb8a5e6945}{%
           family={Hu},
           familyi={H\bibinitperiod},
           given={Bill},
           giveni={B\bibinitperiod}}}%
        {{hash=807ad9155bd748e350fd292c367cc481}{%
           family={Dyer},
           familyi={D\bibinitperiod},
           given={Scott\bibnamedelima Barrett},
           giveni={S\bibinitperiod\bibinitdelim B\bibinitperiod}}}%
      }
      \strng{namehash}{3f211f43cb5f8d4129ccd32acb412bc5}
      \strng{fullhash}{dee54d31e42811ec304a9f877695d038}
      \strng{bibnamehash}{dee54d31e42811ec304a9f877695d038}
      \strng{authorbibnamehash}{dee54d31e42811ec304a9f877695d038}
      \strng{authornamehash}{3f211f43cb5f8d4129ccd32acb412bc5}
      \strng{authorfullhash}{dee54d31e42811ec304a9f877695d038}
      \field{sortinit}{5}
      \field{sortinithash}{20e9b4b0b173788c5dace24730f47d8c}
      \field{labelnamesource}{author}
      \field{labeltitlesource}{title}
      \field{journaltitle}{Sc. Rep.}
      \field{title}{{Long distance seawater intrusion through a karst conduit network in the Woodville Karst Plain, Florida}}
      \field{volume}{6}
      \field{year}{2016}
    \endentry
    \entry{Abbott2014}{article}{}
      \name{author}{1}{}{%
        {{hash=6ad7c0757b2c4f0fdd075f8ec515e5c3}{%
           family={Abbott},
           familyi={A\bibinitperiod},
           given={Alison},
           giveni={A\bibinitperiod}}}%
      }
      \strng{namehash}{6ad7c0757b2c4f0fdd075f8ec515e5c3}
      \strng{fullhash}{6ad7c0757b2c4f0fdd075f8ec515e5c3}
      \strng{bibnamehash}{6ad7c0757b2c4f0fdd075f8ec515e5c3}
      \strng{authorbibnamehash}{6ad7c0757b2c4f0fdd075f8ec515e5c3}
      \strng{authornamehash}{6ad7c0757b2c4f0fdd075f8ec515e5c3}
      \strng{authorfullhash}{6ad7c0757b2c4f0fdd075f8ec515e5c3}
      \field{sortinit}{6}
      \field{sortinithash}{b33bc299efb3c36abec520a4c896a66d}
      \field{labelnamesource}{author}
      \field{labeltitlesource}{title}
      \field{journaltitle}{Nature News}
      \field{month}{5}
      \field{note}{Springer Nature}
      \field{title}{Mexican skeleton gives clue to {American} ancestry}
      \field{year}{2014}
    \endentry
    \entry{macdonald2020paleoindian}{article}{}
      \true{moreauthor}
      \true{morelabelname}
      \name{author}{10}{}{%
        {{hash=1759e7e7a55fb124dcf7b13854fa25ec}{%
           family={MacDonald},
           familyi={M\bibinitperiod},
           given={Brandi\bibnamedelima L},
           giveni={B\bibinitperiod\bibinitdelim L\bibinitperiod}}}%
        {{hash=bf8414d1318ed1c61c8efafd3d51591f}{%
           family={Chatters},
           familyi={C\bibinitperiod},
           given={James\bibnamedelima C},
           giveni={J\bibinitperiod\bibinitdelim C\bibinitperiod}}}%
        {{hash=b1ad1991613afc4460f65dc51899ddf5}{%
           family={Reinhardt},
           familyi={R\bibinitperiod},
           given={Eduard\bibnamedelima G},
           giveni={E\bibinitperiod\bibinitdelim G\bibinitperiod}}}%
        {{hash=e95b4ea24414369252e3f7e890e4a33a}{%
           family={Devos},
           familyi={D\bibinitperiod},
           given={Fred},
           giveni={F\bibinitperiod}}}%
        {{hash=77593f753789a22421f79d6c617d7691}{%
           family={Meacham},
           familyi={M\bibinitperiod},
           given={Sam},
           giveni={S\bibinitperiod}}}%
        {{hash=ab1d6889f236864cd05e14936a7bdbed}{%
           family={Rissolo},
           familyi={R\bibinitperiod},
           given={Dominique},
           giveni={D\bibinitperiod}}}%
        {{hash=8b8e91c88cb6050027e575a17d907f6a}{%
           family={Rock},
           familyi={R\bibinitperiod},
           given={Barry},
           giveni={B\bibinitperiod}}}%
        {{hash=4a4c05e0cbe68373c1812a48eb8bd4d8}{%
           family={Le\bibnamedelima Maillot},
           familyi={L\bibinitperiod\bibinitdelim M\bibinitperiod},
           given={Chris},
           giveni={C\bibinitperiod}}}%
        {{hash=ad76005fe29e3596d3638f21c8544df3}{%
           family={Stalla},
           familyi={S\bibinitperiod},
           given={David},
           giveni={D\bibinitperiod}}}%
        {{hash=1596692bc6c6f50d8c5d6613503b489b}{%
           family={Marino},
           familyi={M\bibinitperiod},
           given={Marc\bibnamedelima D},
           giveni={M\bibinitperiod\bibinitdelim D\bibinitperiod}}}%
      }
      \list{publisher}{1}{%
        {American Association for the Advancement of Science}%
      }
      \strng{namehash}{35ad77b05142882ff58fb16149127707}
      \strng{fullhash}{c20b8b30191e916701cff71a9e38c588}
      \strng{bibnamehash}{c20b8b30191e916701cff71a9e38c588}
      \strng{authorbibnamehash}{c20b8b30191e916701cff71a9e38c588}
      \strng{authornamehash}{35ad77b05142882ff58fb16149127707}
      \strng{authorfullhash}{c20b8b30191e916701cff71a9e38c588}
      \field{sortinit}{6}
      \field{sortinithash}{b33bc299efb3c36abec520a4c896a66d}
      \field{labelnamesource}{author}
      \field{labeltitlesource}{title}
      \field{journaltitle}{Science advances}
      \field{number}{27}
      \field{title}{Paleoindian ochre mines in the submerged caves of the Yucatán Peninsula, Quintana Roo, Mexico}
      \field{volume}{6}
      \field{year}{2020}
      \field{pages}{eaba1219}
      \range{pages}{-1}
    \endentry
    \entry{rissolo2015novel}{inproceedings}{}
      \name{author}{7}{}{%
        {{hash=ab1d6889f236864cd05e14936a7bdbed}{%
           family={Rissolo},
           familyi={R\bibinitperiod},
           given={Dominique},
           giveni={D\bibinitperiod}}}%
        {{hash=a7216c15bc6c2267e5d0e9c407365985}{%
           family={Blank},
           familyi={B\bibinitperiod},
           given={Alberto\bibnamedelima Nava},
           giveni={A\bibinitperiod\bibinitdelim N\bibinitperiod}}}%
        {{hash=b5e09d5dd3046569cb517cc19705b112}{%
           family={Petrovic},
           familyi={P\bibinitperiod},
           given={Vid},
           giveni={V\bibinitperiod}}}%
        {{hash=80ee4208d6e10155ee1d5b85703da37d}{%
           family={Arce},
           familyi={A\bibinitperiod},
           given={Roberto\bibnamedelima Chávez},
           giveni={R\bibinitperiod\bibinitdelim C\bibinitperiod}}}%
        {{hash=be01dd206ea456be27427082739bedda}{%
           family={Jaskolski},
           familyi={J\bibinitperiod},
           given={Corey},
           giveni={C\bibinitperiod}}}%
        {{hash=ed999e6b74d29c3053ed367b4d9efbc2}{%
           family={Erreguerena},
           familyi={E\bibinitperiod},
           given={Pilar\bibnamedelima Luna},
           giveni={P\bibinitperiod\bibinitdelim L\bibinitperiod}}}%
        {{hash=bf8414d1318ed1c61c8efafd3d51591f}{%
           family={Chatters},
           familyi={C\bibinitperiod},
           given={James\bibnamedelima C},
           giveni={J\bibinitperiod\bibinitdelim C\bibinitperiod}}}%
      }
      \list{organization}{1}{%
        {IEEE}%
      }
      \strng{namehash}{a5870abc0bc3595597657dd11db6b254}
      \strng{fullhash}{d646be63cd125299a6ae4a561e24953d}
      \strng{bibnamehash}{d646be63cd125299a6ae4a561e24953d}
      \strng{authorbibnamehash}{d646be63cd125299a6ae4a561e24953d}
      \strng{authornamehash}{a5870abc0bc3595597657dd11db6b254}
      \strng{authorfullhash}{d646be63cd125299a6ae4a561e24953d}
      \field{sortinit}{6}
      \field{sortinithash}{b33bc299efb3c36abec520a4c896a66d}
      \field{labelnamesource}{author}
      \field{labeltitlesource}{title}
      \field{booktitle}{2015 Digital Heritage}
      \field{title}{Novel application of 3D documentation techniques at a submerged Late Pleistocene cave site in Quintana Roo, Mexico}
      \field{volume}{1}
      \field{year}{2015}
      \field{pages}{181\bibrangedash 182}
      \range{pages}{2}
    \endentry
    \entry{de2015ancient}{article}{}
      \name{author}{6}{}{%
        {{hash=7c79710ffe6b8ecb5d7d47094b4d3543}{%
           family={Azevedo},
           familyi={A\bibinitperiod},
           given={Soledad},
           giveni={S\bibinitperiod},
           prefix={de},
           prefixi={d\bibinitperiod}}}%
        {{hash=d1e7de2e508135b8e62c7d2ac7a372ee}{%
           family={Bortolini},
           familyi={B\bibinitperiod},
           given={Maria\bibnamedelima C},
           giveni={M\bibinitperiod\bibinitdelim C\bibinitperiod}}}%
        {{hash=851fe1548be68b9a1ba204f197df4c0a}{%
           family={Bonatto},
           familyi={B\bibinitperiod},
           given={Sandro\bibnamedelima L},
           giveni={S\bibinitperiod\bibinitdelim L\bibinitperiod}}}%
        {{hash=b2ba724c1c80d30eff754184d78e1554}{%
           family={Hünemeier},
           familyi={H\bibinitperiod},
           given={Tábita},
           giveni={T\bibinitperiod}}}%
        {{hash=36e7aa5aacc5af4c0cf01bfab63b5798}{%
           family={Santos},
           familyi={S\bibinitperiod},
           given={Fabrı́cio\bibnamedelima R},
           giveni={F\bibinitperiod\bibinitdelim R\bibinitperiod}}}%
        {{hash=5df8a9835eafdd46b92516e1cabeaecb}{%
           family={González-José},
           familyi={G\bibinithyphendelim J\bibinitperiod},
           given={Rolando},
           giveni={R\bibinitperiod}}}%
      }
      \list{publisher}{1}{%
        {Wiley Online Library}%
      }
      \strng{namehash}{28e7fbdbdbfd4769edae814815433e7e}
      \strng{fullhash}{e6148d3bb1616d7223ea722ad43b7311}
      \strng{bibnamehash}{e6148d3bb1616d7223ea722ad43b7311}
      \strng{authorbibnamehash}{e6148d3bb1616d7223ea722ad43b7311}
      \strng{authornamehash}{28e7fbdbdbfd4769edae814815433e7e}
      \strng{authorfullhash}{e6148d3bb1616d7223ea722ad43b7311}
      \field{sortinit}{6}
      \field{sortinithash}{b33bc299efb3c36abec520a4c896a66d}
      \field{labelnamesource}{author}
      \field{labeltitlesource}{title}
      \field{journaltitle}{American Journal of Physical Anthropology}
      \field{number}{3}
      \field{title}{Ancient remains and the first peopling of the Americas: Reassessing the Hoyo Negro skull}
      \field{volume}{158}
      \field{year}{2015}
      \field{pages}{514\bibrangedash 521}
      \range{pages}{8}
    \endentry
    \entry{KresicMikszewski2013}{book}{}
      \name{author}{2}{}{%
        {{hash=4ead8b72ebe3829d506fb94bf19aec92}{%
           family={Kresic},
           familyi={K\bibinitperiod},
           given={N.},
           giveni={N\bibinitperiod}}}%
        {{hash=edd70b324722964d4dc89f97fc99428a}{%
           family={Mikszewski},
           familyi={M\bibinitperiod},
           given={A.},
           giveni={A\bibinitperiod}}}%
      }
      \list{location}{1}{%
        {Boca Raton, LA}%
      }
      \list{publisher}{1}{%
        {CRC Press}%
      }
      \strng{namehash}{ebfea0c983edc7e254c0795014e32527}
      \strng{fullhash}{ebfea0c983edc7e254c0795014e32527}
      \strng{bibnamehash}{ebfea0c983edc7e254c0795014e32527}
      \strng{authorbibnamehash}{ebfea0c983edc7e254c0795014e32527}
      \strng{authornamehash}{ebfea0c983edc7e254c0795014e32527}
      \strng{authorfullhash}{ebfea0c983edc7e254c0795014e32527}
      \field{sortinit}{7}
      \field{sortinithash}{108d0be1b1bee9773a1173443802c0a3}
      \field{labelnamesource}{author}
      \field{labeltitlesource}{title}
      \field{title}{{Hydrogeological Conceptual Site Models: Data Analysis and Visualization}}
      \field{year}{2013}
    \endentry
    \entry{JoshiAUV2022}{inproceedings}{}
      \name{author}{7}{}{%
        {{hash=cb391d0cc86169b1c74f00d726d5f792}{%
           family={Joshi},
           familyi={J\bibinitperiod},
           given={Bharat},
           giveni={B\bibinitperiod}}}%
        {{hash=152da427b4b46b06285c920f2a579d09}{%
           family={Xanthidis},
           familyi={X\bibinitperiod},
           given={Marios},
           giveni={M\bibinitperiod}}}%
        {{hash=b6ed9ca4bd1a0b4a71b4d83c3333e0f3}{%
           family={Roznere},
           familyi={R\bibinitperiod},
           given={Monika},
           giveni={M\bibinitperiod}}}%
        {{hash=4e393ec74e97736f81c86f26bbf4c0f1}{%
           family={Burgdorfer},
           familyi={B\bibinitperiod},
           given={Nathaniel\bibnamedelima J.},
           giveni={N\bibinitperiod\bibinitdelim J\bibinitperiod}}}%
        {{hash=02570e9fa3d9458928cb75719efd52c0}{%
           family={Mordohai},
           familyi={M\bibinitperiod},
           given={Philippos},
           giveni={P\bibinitperiod}}}%
        {{hash=ee65f89db9f4b051df2847fb0a3f2ecf}{%
           family={Li},
           familyi={L\bibinitperiod},
           given={Alberto\bibnamedelima Quattrini},
           giveni={A\bibinitperiod\bibinitdelim Q\bibinitperiod}}}%
        {{hash=b47c2dc8dc3c61fadae36fa40955b725}{%
           family={Rekleitis},
           familyi={R\bibinitperiod},
           given={Ioannis},
           giveni={I\bibinitperiod}}}%
      }
      \list{location}{1}{%
        {Singapore}%
      }
      \strng{namehash}{80706431c0deda146f392fca79a3395a}
      \strng{fullhash}{f5b520f9c206f8f988ba1cbb3f515872}
      \strng{bibnamehash}{f5b520f9c206f8f988ba1cbb3f515872}
      \strng{authorbibnamehash}{f5b520f9c206f8f988ba1cbb3f515872}
      \strng{authornamehash}{80706431c0deda146f392fca79a3395a}
      \strng{authorfullhash}{f5b520f9c206f8f988ba1cbb3f515872}
      \field{extraname}{1}
      \field{sortinit}{8}
      \field{sortinithash}{a231b008ebf0ecbe0b4d96dcc159445f}
      \field{labelnamesource}{author}
      \field{labeltitlesource}{title}
      \field{abstract}{This paper analyzes the open challenges of exploring and mapping in the underwater realm with the goal of identifying research opportunities that will enable an Autonomous Underwater Vehicle (AUV) to robustly explore different environments. A taxonomy of environments based on their 3D structure is presented together with an analysis on how that influences the camera placement. The difference between exploration and coverage is presented and how they dictate different motion strategies. Loop closure, while critical for the accuracy of the resulting map, proves to be particularly challenging due to the limited field of view and the sensitivity to viewing direction. Experimental results of enforcing loop closures in underwater caves demonstrate a novel navigation strategy. Dense 3D mapping, both online and offline, as well as other sensor configurations are discussed following the presented taxonomy. Experimental results from field trials illustrate the above analysis.}
      \field{booktitle}{IEEE OES AUV Symposium}
      \field{label}{N31}
      \field{month}{9}
      \field{title}{Underwater Exploration and Mapping}
      \field{year}{2022}
      \field{pages}{1\bibrangedash 7}
      \range{pages}{7}
      \verb{file}
      \verb JoshiAUV2022.pdf
      \endverb
    \endentry
    \entry{colmap_sfm}{inproceedings}{}
      \name{author}{2}{}{%
        {{hash=e4ed718dd19b3fee6b128ed86216cb38}{%
           family={Schönberger},
           familyi={S\bibinitperiod},
           given={Johannes\bibnamedelima Lutz},
           giveni={J\bibinitperiod\bibinitdelim L\bibinitperiod}}}%
        {{hash=4af0c62bceb93c143845555052179bc8}{%
           family={Frahm},
           familyi={F\bibinitperiod},
           given={Jan-Michael},
           giveni={J\bibinithyphendelim M\bibinitperiod}}}%
      }
      \strng{namehash}{3a4b1342454668721b1324a50fbe5111}
      \strng{fullhash}{3a4b1342454668721b1324a50fbe5111}
      \strng{bibnamehash}{3a4b1342454668721b1324a50fbe5111}
      \strng{authorbibnamehash}{3a4b1342454668721b1324a50fbe5111}
      \strng{authornamehash}{3a4b1342454668721b1324a50fbe5111}
      \strng{authorfullhash}{3a4b1342454668721b1324a50fbe5111}
      \field{sortinit}{9}
      \field{sortinithash}{0a5ebc79d83c96b6579069544c73c7d4}
      \field{labelnamesource}{author}
      \field{labeltitlesource}{title}
      \field{booktitle}{Conference on Computer Vision and Pattern Recognition (CVPR)}
      \field{title}{Structure-from-Motion Revisited}
      \field{year}{2016}
    \endentry
    \entry{QuattriniLiISERVO2016}{inproceedings}{}
      \name{author}{11}{}{%
        {{hash=ca5720cb4fb280292a8f661739855460}{%
           family={{Quattrini Li}},
           familyi={Q\bibinitperiod},
           given={Alberto},
           giveni={A\bibinitperiod}}}%
        {{hash=7b26cf48524ac59d9b8cc318bd1278c8}{%
           family={Coskun},
           familyi={C\bibinitperiod},
           given={Adem},
           giveni={A\bibinitperiod}}}%
        {{hash=94899ff770ac5865b6cb55654a20aa6d}{%
           family={Doherty},
           familyi={D\bibinitperiod},
           given={Sean\bibnamedelima M.},
           giveni={S\bibinitperiod\bibinitdelim M\bibinitperiod}}}%
        {{hash=f36a1b4c6eda2993777ca363f1804492}{%
           family={Ghasemlou},
           familyi={G\bibinitperiod},
           given={Shervin},
           giveni={S\bibinitperiod}}}%
        {{hash=8f11c57b85277ec54c4a7522afb4f67d}{%
           family={Jagtap},
           familyi={J\bibinitperiod},
           given={Apoorv\bibnamedelima S.},
           giveni={A\bibinitperiod\bibinitdelim S\bibinitperiod}}}%
        {{hash=5e76f79f57bd6bff758145d72232eff9}{%
           family={Modasshir},
           familyi={M\bibinitperiod},
           given={MD},
           giveni={M\bibinitperiod}}}%
        {{hash=99f26b0ff5ddff59d005db7acd08100b}{%
           family={Rahman},
           familyi={R\bibinitperiod},
           given={Sharmin},
           giveni={S\bibinitperiod}}}%
        {{hash=275eb008ad8c6b74538e17f29f190d72}{%
           family={Singh},
           familyi={S\bibinitperiod},
           given={Akanksha},
           giveni={A\bibinitperiod}}}%
        {{hash=152da427b4b46b06285c920f2a579d09}{%
           family={Xanthidis},
           familyi={X\bibinitperiod},
           given={Marios},
           giveni={M\bibinitperiod}}}%
        {{hash=62719bed7b95c2b576bf6b784ca15935}{%
           family={O'Kane},
           familyi={O\bibinitperiod},
           given={Jason\bibnamedelima M.},
           giveni={J\bibinitperiod\bibinitdelim M\bibinitperiod}}}%
        {{hash=b47c2dc8dc3c61fadae36fa40955b725}{%
           family={Rekleitis},
           familyi={R\bibinitperiod},
           given={Ioannis},
           giveni={I\bibinitperiod}}}%
      }
      \list{location}{1}{%
        {Tokyo, Japan}%
      }
      \strng{namehash}{98f5116ab597edfbbcb46666f0a3da79}
      \strng{fullhash}{4a5feedb7023b112d932762aeedd3ca4}
      \strng{bibnamehash}{b00fe9e5edb4d51d0729b8f4d03a682c}
      \strng{authorbibnamehash}{b00fe9e5edb4d51d0729b8f4d03a682c}
      \strng{authornamehash}{98f5116ab597edfbbcb46666f0a3da79}
      \strng{authorfullhash}{4a5feedb7023b112d932762aeedd3ca4}
      \field{sortinit}{1}
      \field{sortinithash}{4f6aaa89bab872aa0999fec09ff8e98a}
      \field{labelnamesource}{author}
      \field{labeltitlesource}{title}
      \field{abstract}{The problem of state estimation using primarily visual data has received a lot of attention in the last decade. Several open source packages have appeared addressing the problem, each supported by impressive demonstrations. Applying any of these packages on a new dataset however, has been proven extremely challenging. Suboptimal performance, loss of localization, and challenges in customization have not produced a clear winner. Several other research groups have presented superb performance without releasing the code, sometimes materializing as commercial products. In this paper, ten of the most promising open source packages are evaluated, by cross validating them on the datasets provided for each package and by testing them on eight different datasets collected over the years in our laboratory. Indoor and outdoor, terrestrial and flying vehicles, in addition to underwater robots, cameras, and buoys were used to collect data. An analysis on the motions required for the different approaches and an evaluation of their performance is presented.}
      \field{booktitle}{International Symposium of Experimental Robotics (ISER)}
      \field{label}{C65}
      \field{month}{3}
      \field{title}{Experimental Comparison of open source Vision based State Estimation Algorithms}
      \field{year}{2016}
      \verb{file}
      \verb RekleitisISERVO2016.pdf
      \endverb
    \endentry
    \entry{JoshiIROS2019}{inproceedings}{}
      \name{author}{11}{}{%
        {{hash=cb391d0cc86169b1c74f00d726d5f792}{%
           family={Joshi},
           familyi={J\bibinitperiod},
           given={Bharat},
           giveni={B\bibinitperiod}}}%
        {{hash=99f26b0ff5ddff59d005db7acd08100b}{%
           family={Rahman},
           familyi={R\bibinitperiod},
           given={Sharmin},
           giveni={S\bibinitperiod}}}%
        {{hash=2a7c69d9583ce63c0023765f897a1372}{%
           family={Kalaitzakis},
           familyi={K\bibinitperiod},
           given={Michail},
           giveni={M\bibinitperiod}}}%
        {{hash=53de242f609f44c2e6475b66a63bcda5}{%
           family={Cain},
           familyi={C\bibinitperiod},
           given={Brennan},
           giveni={B\bibinitperiod}}}%
        {{hash=89376d05e0012a14a3fc913258e2d066}{%
           family={Johnson},
           familyi={J\bibinitperiod},
           given={James},
           giveni={J\bibinitperiod}}}%
        {{hash=152da427b4b46b06285c920f2a579d09}{%
           family={Xanthidis},
           familyi={X\bibinitperiod},
           given={Marios},
           giveni={M\bibinitperiod}}}%
        {{hash=c466b011cd605f3f1c67570732be8b67}{%
           family={Karapetyan},
           familyi={K\bibinitperiod},
           given={Nare},
           giveni={N\bibinitperiod}}}%
        {{hash=f5dba97885a2b06908de19694da2fcaa}{%
           family={Hernandez},
           familyi={H\bibinitperiod},
           given={Alan},
           giveni={A\bibinitperiod}}}%
        {{hash=ca5720cb4fb280292a8f661739855460}{%
           family={{Quattrini Li}},
           familyi={Q\bibinitperiod},
           given={Alberto},
           giveni={A\bibinitperiod}}}%
        {{hash=4161e8964577630a2a8070e3c4541d5d}{%
           family={Vitzilaios},
           familyi={V\bibinitperiod},
           given={Nikolaos},
           giveni={N\bibinitperiod}}}%
        {{hash=b47c2dc8dc3c61fadae36fa40955b725}{%
           family={Rekleitis},
           familyi={R\bibinitperiod},
           given={Ioannis},
           giveni={I\bibinitperiod}}}%
      }
      \list{location}{1}{%
        {Macau}%
      }
      \strng{namehash}{80706431c0deda146f392fca79a3395a}
      \strng{fullhash}{f13adc85dd31cf5ce4ff0d6741706eee}
      \strng{bibnamehash}{452aaad4c694a1dd15e35aae0be04b02}
      \strng{authorbibnamehash}{452aaad4c694a1dd15e35aae0be04b02}
      \strng{authornamehash}{80706431c0deda146f392fca79a3395a}
      \strng{authorfullhash}{f13adc85dd31cf5ce4ff0d6741706eee}
      \field{extraname}{2}
      \field{sortinit}{1}
      \field{sortinithash}{4f6aaa89bab872aa0999fec09ff8e98a}
      \field{labelnamesource}{author}
      \field{labeltitlesource}{title}
      \field{abstract}{A plethora of state estimation techniques have appeared in the last decade using visual data, and more recently with added inertial data. Datasets typically used for evaluation include indoor and urban environments, where supporting videos have shown impressive performance. However, such techniques have not been fully evaluated in challenging conditions, such as the marine domain. In this paper, we compare ten recent open-source packages to provide insights on their performance and guidelines on addressing current challenges. Specifically, we selected direct methods and tightly-coupled optimization techniques that fuse camera and Inertial Measurement Unit (IMU) data together. Experiments are conducted by testing all packages on datasets collected over the years with underwater robots in our laboratory. All the datasets are made available online.}
      \field{booktitle}{IEEE/RSJ International Conference on Intelligent Robots and Systems (IROS)}
      \field{label}{C86}
      \field{month}{11}
      \field{title}{{Experimental Comparison of Open Source Visual-Inertial-Based State Estimation Algorithms in the Underwater Domain}}
      \field{year}{2019}
      \field{pages}{7221\bibrangedash 7227}
      \range{pages}{7}
      \verb{doi}
      \verb https://doi.org/10.1109/iros40897.2019.8968049
      \endverb
      \verb{file}
      \verb IROS19_0500_FI.pdf
      \endverb
      \verb{urlraw}
      \verb https://arxiv.org/abs/1904.02215
      \endverb
      \verb{url}
      \verb https://arxiv.org/abs/1904.02215
      \endverb
    \endentry
    \entry{RahmanICRA2018}{inproceedings}{}
      \name{author}{3}{}{%
        {{hash=99f26b0ff5ddff59d005db7acd08100b}{%
           family={Rahman},
           familyi={R\bibinitperiod},
           given={Sharmin},
           giveni={S\bibinitperiod}}}%
        {{hash=ca5720cb4fb280292a8f661739855460}{%
           family={{Quattrini Li}},
           familyi={Q\bibinitperiod},
           given={Alberto},
           giveni={A\bibinitperiod}}}%
        {{hash=b47c2dc8dc3c61fadae36fa40955b725}{%
           family={Rekleitis},
           familyi={R\bibinitperiod},
           given={Ioannis},
           giveni={I\bibinitperiod}}}%
      }
      \list{location}{1}{%
        {Brisbane, Australia}%
      }
      \strng{namehash}{19daca79b3da14b70f0ca0897926824c}
      \strng{fullhash}{5c0d02b90f546f2edf915d34ace91da3}
      \strng{bibnamehash}{5c0d02b90f546f2edf915d34ace91da3}
      \strng{authorbibnamehash}{5c0d02b90f546f2edf915d34ace91da3}
      \strng{authornamehash}{19daca79b3da14b70f0ca0897926824c}
      \strng{authorfullhash}{5c0d02b90f546f2edf915d34ace91da3}
      \field{extraname}{2}
      \field{sortinit}{1}
      \field{sortinithash}{4f6aaa89bab872aa0999fec09ff8e98a}
      \field{labelnamesource}{author}
      \field{labeltitlesource}{title}
      \field{abstract}{This paper presents an extension to a state of the art Visual-Inertial state estimation package (OKVIS) in order to accommodate data from an underwater acoustic sensor. Mapping underwater structures is important in several fields, such as marine archaeology, search and rescue, resource management, hydrogeology, and speleology. Collecting the data, however, is a challenging, dangerous, and exhausting task. The underwater domain presents unique challenges in the quality of the visual data available; as such, augmenting the exteroceptive sensing with acoustic range data results in improved reconstructions of the underwater structures. Experimental results from underwater wrecks, an underwater cave, and a submerged bus demonstrate the performance of our approach.}
      \field{booktitle}{IEEE International Conference on Robotics and Automation}
      \field{label}{C77}
      \field{month}{5}
      \field{title}{{Sonar Visual Inertial SLAM of Underwater Structures}}
      \field{year}{2018}
      \field{pages}{5190\bibrangedash 5196}
      \range{pages}{7}
      \verb{doi}
      \verb https://doi.org/10.1109/icra.2018.8460545
      \endverb
      \verb{file}
      \verb RahmanICRA2018.pdf
      \endverb
    \endentry
    \entry{RahmanIROS2019a}{inproceedings}{}
      \name{author}{3}{}{%
        {{hash=99f26b0ff5ddff59d005db7acd08100b}{%
           family={Rahman},
           familyi={R\bibinitperiod},
           given={Sharmin},
           giveni={S\bibinitperiod}}}%
        {{hash=ca5720cb4fb280292a8f661739855460}{%
           family={{Quattrini Li}},
           familyi={Q\bibinitperiod},
           given={Alberto},
           giveni={A\bibinitperiod}}}%
        {{hash=b47c2dc8dc3c61fadae36fa40955b725}{%
           family={Rekleitis},
           familyi={R\bibinitperiod},
           given={Ioannis},
           giveni={I\bibinitperiod}}}%
      }
      \list{location}{1}{%
        {Macau, (IROS ICROS Best Application Paper Award. Finalist)}%
      }
      \strng{namehash}{19daca79b3da14b70f0ca0897926824c}
      \strng{fullhash}{5c0d02b90f546f2edf915d34ace91da3}
      \strng{bibnamehash}{5c0d02b90f546f2edf915d34ace91da3}
      \strng{authorbibnamehash}{5c0d02b90f546f2edf915d34ace91da3}
      \strng{authornamehash}{19daca79b3da14b70f0ca0897926824c}
      \strng{authorfullhash}{5c0d02b90f546f2edf915d34ace91da3}
      \field{extraname}{3}
      \field{sortinit}{1}
      \field{sortinithash}{4f6aaa89bab872aa0999fec09ff8e98a}
      \field{labelnamesource}{author}
      \field{labeltitlesource}{title}
      \field{abstract}{This paper presents a novel tightly-coupled keyframe-based Simultaneous Localization and Mapping (SLAM) system with loop-closing and relocalization capabilities targeted for the underwater domain. Our previous work, SVIn, augmented the state-of-the-art visual-inertial state estimation package OKVIS to accommodate acoustic data from sonar in a non-linear optimization-based framework. This paper addresses drift and loss of localization – one of the main problems affecting other packages in underwater domain – by providing the following main contributions: a robust initialization method to refine scale using depth measurements, a fast preprocessing step to enhance the image quality, and a real-time loop-closing and relocalization method using bag of words (BoW). An additional contribution is the addition of depth measurements from a pressure sensor to the tightly-coupled optimization formulation. Experimental results on datasets collected with a custom-made underwater sensor suite and an autonomous underwater vehicle from challenging underwater environments with poor visibility demonstrate performance never achieved before in terms of accuracy and robustness.}
      \field{booktitle}{IEEE/RSJ International Conference on Intelligent Robots and Systems (IROS)}
      \field{label}{C85}
      \field{month}{11}
      \field{title}{{An Underwater SLAM System using Sonar, Visual, Inertial, and Depth Sensor}}
      \field{year}{2019}
      \field{pages}{1861\bibrangedash 1868}
      \range{pages}{8}
      \verb{doi}
      \verb https://doi.org/10.1109/iros40897.2019.8967703
      \endverb
      \verb{file}
      \verb IROS19_0269_FI.pdf
      \endverb
    \endentry
    \entry{RahmanIJRR2022}{article}{}
      \name{author}{3}{}{%
        {{hash=99f26b0ff5ddff59d005db7acd08100b}{%
           family={Rahman},
           familyi={R\bibinitperiod},
           given={Sharmin},
           giveni={S\bibinitperiod}}}%
        {{hash=ca5720cb4fb280292a8f661739855460}{%
           family={{Quattrini Li}},
           familyi={Q\bibinitperiod},
           given={Alberto},
           giveni={A\bibinitperiod}}}%
        {{hash=b47c2dc8dc3c61fadae36fa40955b725}{%
           family={Rekleitis},
           familyi={R\bibinitperiod},
           given={Ioannis},
           giveni={I\bibinitperiod}}}%
      }
      \strng{namehash}{19daca79b3da14b70f0ca0897926824c}
      \strng{fullhash}{5c0d02b90f546f2edf915d34ace91da3}
      \strng{bibnamehash}{5c0d02b90f546f2edf915d34ace91da3}
      \strng{authorbibnamehash}{5c0d02b90f546f2edf915d34ace91da3}
      \strng{authornamehash}{19daca79b3da14b70f0ca0897926824c}
      \strng{authorfullhash}{5c0d02b90f546f2edf915d34ace91da3}
      \field{extraname}{4}
      \field{sortinit}{1}
      \field{sortinithash}{4f6aaa89bab872aa0999fec09ff8e98a}
      \field{labelnamesource}{author}
      \field{labeltitlesource}{title}
      \field{abstract}{This paper presents SVIn2, a novel tightly-coupled keyframe-based Simultaneous Localization and Mapping (SLAM) system, which fuses Scanning Profiling Sonar, Visual, Inertial, and water pressure information in a non-linear optimization framework for small and large scale challenging underwater environments. The developed real-time system features robust initialization, loop-closing, and relocalization capabilities, which make the system reliable in the presence of haze, blurriness, low light, and lighting variations, typically observed in underwater scenarios. Over the last decade, Visual-Inertial Odometry (VIO) and SLAM systems have shown excellent performance for mobile robots in indoor and outdoor environments, but often fail underwater due to the inherent difficulties in such environments. Our approach combats the weaknesses of previous approaches by utilizing additional sensors and exploiting their complementary characteristics. In particular, we use (1) acoustic range information for improved reconstruction and localization, thanks to the reliable distance measurement; (2) depth information from water pressure sensor for robust initialization, refining the scale, and assisting to limit the drift in the tightly-coupled integration. The developed software -- made open source -- has been successfully used to test and validate the proposed system in both benchmark datasets and numerous real world underwater scenarios, including datasets collected with a custom-made underwater sensor suite and an autonomous underwater vehicle (AUV) Aqua2. SVIn2 demonstrated outstanding performance in terms of accuracy and robustness on those datasets and enabled other robotic tasks, e.g., planning for underwater robots in presence of obstacles.}
      \field{journaltitle}{International Journal of Robotics Research}
      \field{label}{J17}
      \field{month}{07}
      \field{number}{11-12}
      \field{title}{{SVIn2: A Multi-sensor Fusion-based Underwater SLAM System}}
      \field{volume}{41}
      \field{year}{2022}
      \field{pages}{1022\bibrangedash 1042}
      \range{pages}{21}
      \verb{doi}
      \verb 10.1177/02783649221110259
      \endverb
      \verb{file}
      \verb RahmanIJRR2022.pdf
      \endverb
    \endentry
    \entry{JoshiICRA2022}{inproceedings}{}
      \name{author}{4}{}{%
        {{hash=cb391d0cc86169b1c74f00d726d5f792}{%
           family={Joshi},
           familyi={J\bibinitperiod},
           given={Bharat},
           giveni={B\bibinitperiod}}}%
        {{hash=152da427b4b46b06285c920f2a579d09}{%
           family={Xanthidis},
           familyi={X\bibinitperiod},
           given={Marios},
           giveni={M\bibinitperiod}}}%
        {{hash=99f26b0ff5ddff59d005db7acd08100b}{%
           family={Rahman},
           familyi={R\bibinitperiod},
           given={Sharmin},
           giveni={S\bibinitperiod}}}%
        {{hash=b47c2dc8dc3c61fadae36fa40955b725}{%
           family={Rekleitis},
           familyi={R\bibinitperiod},
           given={Ioannis},
           giveni={I\bibinitperiod}}}%
      }
      \list{location}{1}{%
        {Philadelphia, PA, USA}%
      }
      \strng{namehash}{80706431c0deda146f392fca79a3395a}
      \strng{fullhash}{459a38e198aa41b1d826031490f4b84a}
      \strng{bibnamehash}{459a38e198aa41b1d826031490f4b84a}
      \strng{authorbibnamehash}{459a38e198aa41b1d826031490f4b84a}
      \strng{authornamehash}{80706431c0deda146f392fca79a3395a}
      \strng{authorfullhash}{459a38e198aa41b1d826031490f4b84a}
      \field{extraname}{3}
      \field{sortinit}{1}
      \field{sortinithash}{4f6aaa89bab872aa0999fec09ff8e98a}
      \field{labelnamesource}{author}
      \field{labeltitlesource}{title}
      \field{abstract}{In this paper we present a complete framework for Underwater SLAM utilizing a single inexpensive sensor. Over the recent years, imaging technology of action cameras is producing	stunning results even under the challenging conditions of the underwater domain. The GoPro 9 camera provides high definition video in synchronization with an Inertial Measurement Unit (IMU) data stream encoded in a single mp4 file. The visual inertial SLAM framework is augmented to adjust the map after each loop closure. Data collected at an artificial wreck of the coast	of South Carolina and in caverns and caves in Florida demonstrate the robustness of the proposed approach in a variety of conditions.}
      \field{booktitle}{IEEE International Conference on Robotics and Automation (ICRA)}
      \field{label}{C93}
      \field{title}{High Definition, Inexpensive, Underwater Mapping}
      \field{year}{2022}
      \field{pages}{1113\bibrangedash 1121}
      \range{pages}{9}
      \verb{file}
      \verb ICRA22_1224_FI.pdf
      \endverb
    \endentry
    \entry{orbslam3}{article}{}
      \name{author}{5}{}{%
        {{hash=6cdacf84be87337589747b5b4820bfba}{%
           family={Campos},
           familyi={C\bibinitperiod},
           given={Carlos},
           giveni={C\bibinitperiod}}}%
        {{hash=45619e807fa5241f5da5135ce7992309}{%
           family={Elvira},
           familyi={E\bibinitperiod},
           given={Richard},
           giveni={R\bibinitperiod}}}%
        {{hash=9f43a0aba53c88080ad5450b521f709d}{%
           family={Rodríguez},
           familyi={R\bibinitperiod},
           given={Juan\bibnamedelimb J.\bibnamedelimi Gómez},
           giveni={J\bibinitperiod\bibinitdelim J\bibinitperiod\bibinitdelim G\bibinitperiod}}}%
        {{hash=adb8e8b8de517dfbf22d3d4fd7966ac0}{%
           family={M.\bibnamedelimi Montiel},
           familyi={M\bibinitperiod\bibinitdelim M\bibinitperiod},
           given={José\bibnamedelima M.},
           giveni={J\bibinitperiod\bibinitdelim M\bibinitperiod}}}%
        {{hash=131e28fbc8f021936ec0200f8418acbd}{%
           family={D.\bibnamedelimi Tardós},
           familyi={D\bibinitperiod\bibinitdelim T\bibinitperiod},
           given={Juan},
           giveni={J\bibinitperiod}}}%
      }
      \strng{namehash}{4b5f61bd95eca9030bea99344bcf2557}
      \strng{fullhash}{f9f6d93fdb2be95c807f7bfffac0e594}
      \strng{bibnamehash}{f9f6d93fdb2be95c807f7bfffac0e594}
      \strng{authorbibnamehash}{f9f6d93fdb2be95c807f7bfffac0e594}
      \strng{authornamehash}{4b5f61bd95eca9030bea99344bcf2557}
      \strng{authorfullhash}{f9f6d93fdb2be95c807f7bfffac0e594}
      \field{sortinit}{1}
      \field{sortinithash}{4f6aaa89bab872aa0999fec09ff8e98a}
      \field{labelnamesource}{author}
      \field{labeltitlesource}{title}
      \field{journaltitle}{IEEE Transactions on Robotics}
      \field{title}{{ORB-SLAM3}: An Accurate Open-Source Library for Visual, Visual–Inertial, and Multimap SLAM}
      \field{year}{2021}
    \endentry
    \entry{geneva2020openvins}{inproceedings}{}
      \name{author}{5}{}{%
        {{hash=78bc90225e42455a0bbd25dd3a8a29d1}{%
           family={Geneva},
           familyi={G\bibinitperiod},
           given={Patrick},
           giveni={P\bibinitperiod}}}%
        {{hash=f12c7709fd827b53d8fbd7272af059be}{%
           family={Eckenhoff},
           familyi={E\bibinitperiod},
           given={Kevin},
           giveni={K\bibinitperiod}}}%
        {{hash=61bcbd27eb2f2ec56b68883ac8681e40}{%
           family={Lee},
           familyi={L\bibinitperiod},
           given={Woosik},
           giveni={W\bibinitperiod}}}%
        {{hash=95308a058f3d04da8f063db1b18a273e}{%
           family={Yang},
           familyi={Y\bibinitperiod},
           given={Yulin},
           giveni={Y\bibinitperiod}}}%
        {{hash=7c5de65888971006a35030ffb0466a50}{%
           family={Huang},
           familyi={H\bibinitperiod},
           given={Guoquan},
           giveni={G\bibinitperiod}}}%
      }
      \list{organization}{1}{%
        {IEEE}%
      }
      \strng{namehash}{4e5c481409e00e589f9e20a939c10c8c}
      \strng{fullhash}{2d2574ea89f9a0bee1e323922da8b9fb}
      \strng{bibnamehash}{2d2574ea89f9a0bee1e323922da8b9fb}
      \strng{authorbibnamehash}{2d2574ea89f9a0bee1e323922da8b9fb}
      \strng{authornamehash}{4e5c481409e00e589f9e20a939c10c8c}
      \strng{authorfullhash}{2d2574ea89f9a0bee1e323922da8b9fb}
      \field{sortinit}{1}
      \field{sortinithash}{4f6aaa89bab872aa0999fec09ff8e98a}
      \field{labelnamesource}{author}
      \field{labeltitlesource}{title}
      \field{booktitle}{2020 IEEE International Conference on Robotics and Automation (ICRA)}
      \field{title}{Openvins: A research platform for visual-inertial estimation}
      \field{year}{2020}
      \field{pages}{4666\bibrangedash 4672}
      \range{pages}{7}
    \endentry
    \entry{madwick_filter}{inproceedings}{}
      \name{author}{3}{}{%
        {{hash=907c1fecd31054023a0459215498ad74}{%
           family={Madgwick},
           familyi={M\bibinitperiod},
           given={Sebastian\bibnamedelimb O.\bibnamedelimi H.},
           giveni={S\bibinitperiod\bibinitdelim O\bibinitperiod\bibinitdelim H\bibinitperiod}}}%
        {{hash=c69bf9c458890c8420c2873ead4dc352}{%
           family={Harrison},
           familyi={H\bibinitperiod},
           given={Andrew\bibnamedelimb J.\bibnamedelimi L.},
           giveni={A\bibinitperiod\bibinitdelim J\bibinitperiod\bibinitdelim L\bibinitperiod}}}%
        {{hash=8386d6fdb31b7d08180eba6d9b36c82a}{%
           family={Vaidyanathan},
           familyi={V\bibinitperiod},
           given={Ravi},
           giveni={R\bibinitperiod}}}%
      }
      \strng{namehash}{a02f4c1839f52030ec0bf0d418296c9e}
      \strng{fullhash}{c5811219bd0bac5e88358a71c6bf74f9}
      \strng{bibnamehash}{c5811219bd0bac5e88358a71c6bf74f9}
      \strng{authorbibnamehash}{c5811219bd0bac5e88358a71c6bf74f9}
      \strng{authornamehash}{a02f4c1839f52030ec0bf0d418296c9e}
      \strng{authorfullhash}{c5811219bd0bac5e88358a71c6bf74f9}
      \field{sortinit}{1}
      \field{sortinithash}{4f6aaa89bab872aa0999fec09ff8e98a}
      \field{labelnamesource}{author}
      \field{labeltitlesource}{title}
      \field{booktitle}{2011 IEEE International Conference on Rehabilitation Robotics}
      \field{title}{Estimation of IMU and MARG orientation using a gradient descent algorithm}
      \field{year}{2011}
      \field{pages}{1\bibrangedash 7}
      \range{pages}{7}
      \verb{doi}
      \verb 10.1109/ICORR.2011.5975346
      \endverb
    \endentry
    \entry{complimentary_filter}{article}{}
      \name{author}{3}{}{%
        {{hash=79fe9c73cf187803e4113139a8bcb5b2}{%
           family={Valenti},
           familyi={V\bibinitperiod},
           given={Roberto\bibnamedelima G.},
           giveni={R\bibinitperiod\bibinitdelim G\bibinitperiod}}}%
        {{hash=03a32d5debc65eaa87a838a52b0232d2}{%
           family={Dryanovski},
           familyi={D\bibinitperiod},
           given={Ivan},
           giveni={I\bibinitperiod}}}%
        {{hash=8175883d96517eae0a70fca6859fff9c}{%
           family={Xiao},
           familyi={X\bibinitperiod},
           given={Jizhong},
           giveni={J\bibinitperiod}}}%
      }
      \strng{namehash}{e4ee3c140ce3c3c40c6c1ee375575586}
      \strng{fullhash}{2a45ea8b4e1dfaf4b8e22e218cd8140b}
      \strng{bibnamehash}{2a45ea8b4e1dfaf4b8e22e218cd8140b}
      \strng{authorbibnamehash}{2a45ea8b4e1dfaf4b8e22e218cd8140b}
      \strng{authornamehash}{e4ee3c140ce3c3c40c6c1ee375575586}
      \strng{authorfullhash}{2a45ea8b4e1dfaf4b8e22e218cd8140b}
      \field{sortinit}{1}
      \field{sortinithash}{4f6aaa89bab872aa0999fec09ff8e98a}
      \field{labelnamesource}{author}
      \field{labeltitlesource}{title}
      \field{issn}{1424-8220}
      \field{journaltitle}{Sensors}
      \field{number}{8}
      \field{title}{Keeping a Good Attitude: A Quaternion-Based Orientation Filter for IMUs and MARGs}
      \field{volume}{15}
      \field{year}{2015}
      \field{pages}{19302\bibrangedash 19330}
      \range{pages}{29}
      \verb{doi}
      \verb 10.3390/s150819302
      \endverb
      \verb{urlraw}
      \verb https://www.mdpi.com/1424-8220/15/8/19302
      \endverb
      \verb{url}
      \verb https://www.mdpi.com/1424-8220/15/8/19302
      \endverb
    \endentry
    \entry{zhao_imu_mag_ultrasonic}{article}{}
      \name{author}{2}{}{%
        {{hash=f020e3fcb5d7b3ca19b58f57e8a2b4f3}{%
           family={Zhao},
           familyi={Z\bibinitperiod},
           given={He},
           giveni={H\bibinitperiod}}}%
        {{hash=fa469f9e9b78a84bd715f50f0971b536}{%
           family={Wang},
           familyi={W\bibinitperiod},
           given={Zheyao},
           giveni={Z\bibinitperiod}}}%
      }
      \strng{namehash}{bd4bb37cb707b6012dc9c47cec569434}
      \strng{fullhash}{bd4bb37cb707b6012dc9c47cec569434}
      \strng{bibnamehash}{bd4bb37cb707b6012dc9c47cec569434}
      \strng{authorbibnamehash}{bd4bb37cb707b6012dc9c47cec569434}
      \strng{authornamehash}{bd4bb37cb707b6012dc9c47cec569434}
      \strng{authorfullhash}{bd4bb37cb707b6012dc9c47cec569434}
      \field{sortinit}{1}
      \field{sortinithash}{4f6aaa89bab872aa0999fec09ff8e98a}
      \field{labelnamesource}{author}
      \field{labeltitlesource}{title}
      \field{journaltitle}{IEEE Sensors Journal}
      \field{number}{5}
      \field{title}{Motion Measurement Using Inertial Sensors, Ultrasonic Sensors, and Magnetometers With Extended Kalman Filter for Data Fusion}
      \field{volume}{12}
      \field{year}{2012}
      \field{pages}{943\bibrangedash 953}
      \range{pages}{11}
      \verb{doi}
      \verb 10.1109/JSEN.2011.2166066
      \endverb
    \endentry
    \entry{wang_vimo}{article}{}
      \name{author}{5}{}{%
        {{hash=0a303576467220cc9a4008b3ab65f6a0}{%
           family={Wang},
           familyi={W\bibinitperiod},
           given={Jingzhe},
           giveni={J\bibinitperiod}}}%
        {{hash=79621144453d829a04b5210f69a2e445}{%
           family={Li},
           familyi={L\bibinitperiod},
           given={Leilei},
           giveni={L\bibinitperiod}}}%
        {{hash=e0aec565917f246fae024b0cc4790d5a}{%
           family={Yu},
           familyi={Y\bibinitperiod},
           given={Huan},
           giveni={H\bibinitperiod}}}%
        {{hash=bc09ed1e0265efd158ce781f89c88452}{%
           family={Gui},
           familyi={G\bibinitperiod},
           given={Xunya},
           giveni={X\bibinitperiod}}}%
        {{hash=a4899033116a5d6cd24901cfbb004d69}{%
           family={Li},
           familyi={L\bibinitperiod},
           given={Zucheng},
           giveni={Z\bibinitperiod}}}%
      }
      \strng{namehash}{2331896483611df896e644447990cdd8}
      \strng{fullhash}{94506ee735209349813d95868b5c642f}
      \strng{bibnamehash}{94506ee735209349813d95868b5c642f}
      \strng{authorbibnamehash}{94506ee735209349813d95868b5c642f}
      \strng{authornamehash}{2331896483611df896e644447990cdd8}
      \strng{authorfullhash}{94506ee735209349813d95868b5c642f}
      \field{sortinit}{1}
      \field{sortinithash}{4f6aaa89bab872aa0999fec09ff8e98a}
      \field{labelnamesource}{author}
      \field{labeltitlesource}{title}
      \field{issn}{1424-8220}
      \field{journaltitle}{Sensors}
      \field{number}{16}
      \field{title}{VIMO: A Visual-Inertial-Magnetic Navigation System Based on Non-Linear Optimization}
      \field{volume}{20}
      \field{year}{2020}
      \verb{urlraw}
      \verb https://www.mdpi.com/1424-8220/20/16/4386
      \endverb
      \verb{url}
      \verb https://www.mdpi.com/1424-8220/20/16/4386
      \endverb
    \endentry
    \entry{Siebler}{article}{}
      \name{author}{3}{}{%
        {{hash=5318b743078945b6b7a0a4d820f41845}{%
           family={Siebler},
           familyi={S\bibinitperiod},
           given={Benjamin},
           giveni={B\bibinitperiod}}}%
        {{hash=8bf107a2f05747ae4d366f50266dfcc6}{%
           family={Sand},
           familyi={S\bibinitperiod},
           given={Stephan},
           giveni={S\bibinitperiod}}}%
        {{hash=5520e2275056a55170c6896464012581}{%
           family={Hanebeck},
           familyi={H\bibinitperiod},
           given={Uwe\bibnamedelima D.},
           giveni={U\bibinitperiod\bibinitdelim D\bibinitperiod}}}%
      }
      \strng{namehash}{18575034242363c384bfe608327a8cca}
      \strng{fullhash}{c1e3b534694b554ed5a234e54e421b66}
      \strng{bibnamehash}{c1e3b534694b554ed5a234e54e421b66}
      \strng{authorbibnamehash}{c1e3b534694b554ed5a234e54e421b66}
      \strng{authornamehash}{18575034242363c384bfe608327a8cca}
      \strng{authorfullhash}{c1e3b534694b554ed5a234e54e421b66}
      \field{sortinit}{1}
      \field{sortinithash}{4f6aaa89bab872aa0999fec09ff8e98a}
      \field{labelnamesource}{author}
      \field{labeltitlesource}{title}
      \field{journaltitle}{IEEE Sensors Journal}
      \field{number}{3}
      \field{title}{Localization With Magnetic Field Distortions and Simultaneous Magnetometer Calibration}
      \field{volume}{21}
      \field{year}{2021}
      \field{pages}{3388\bibrangedash 3397}
      \range{pages}{10}
      \verb{doi}
      \verb 10.1109/JSEN.2020.3024073
      \endverb
    \endentry
    \entry{caruso_mimu}{inproceedings}{}
      \name{author}{5}{}{%
        {{hash=1cd216229e2fb0072c5a26fa6837024e}{%
           family={Caruso},
           familyi={C\bibinitperiod},
           given={David},
           giveni={D\bibinitperiod}}}%
        {{hash=0f05fd141ef08a698be3f3d0f0878a96}{%
           family={Eudes},
           familyi={E\bibinitperiod},
           given={Alexandre},
           giveni={A\bibinitperiod}}}%
        {{hash=313d18a818fc7479f0606d72d5b95b81}{%
           family={Sanfourche},
           familyi={S\bibinitperiod},
           given={Martial},
           giveni={M\bibinitperiod}}}%
        {{hash=1c8e82fd8e4ceffffd1f8e982ad0db1a}{%
           family={Vissière},
           familyi={V\bibinitperiod},
           given={David},
           giveni={D\bibinitperiod}}}%
        {{hash=b1056fb64ffd8b569084cd51ad8d2c4a}{%
           family={Besnerais},
           familyi={B\bibinitperiod},
           given={Guy},
           giveni={G\bibinitperiod},
           prefix={le},
           prefixi={l\bibinitperiod}}}%
      }
      \strng{namehash}{157b54281ec7f717960ab21c843a9e5a}
      \strng{fullhash}{be963d74a300667257c45fd7348da7cd}
      \strng{bibnamehash}{be963d74a300667257c45fd7348da7cd}
      \strng{authorbibnamehash}{be963d74a300667257c45fd7348da7cd}
      \strng{authornamehash}{157b54281ec7f717960ab21c843a9e5a}
      \strng{authorfullhash}{be963d74a300667257c45fd7348da7cd}
      \field{extraname}{1}
      \field{sortinit}{2}
      \field{sortinithash}{8b555b3791beccb63322c22f3320aa9a}
      \field{labelnamesource}{author}
      \field{labeltitlesource}{title}
      \field{booktitle}{2017 IEEE/RSJ International Conference on Intelligent Robots and Systems (IROS)}
      \field{title}{Robust indoor/outdoor navigation through magneto-visual-inertial optimization-based estimation}
      \field{year}{2017}
      \field{pages}{4402\bibrangedash 4409}
      \range{pages}{8}
      \verb{doi}
      \verb 10.1109/IROS.2017.8206304
      \endverb
    \endentry
    \entry{caruso_mimu_kf}{inproceedings}{}
      \name{author}{5}{}{%
        {{hash=1cd216229e2fb0072c5a26fa6837024e}{%
           family={Caruso},
           familyi={C\bibinitperiod},
           given={David},
           giveni={D\bibinitperiod}}}%
        {{hash=0f05fd141ef08a698be3f3d0f0878a96}{%
           family={Eudes},
           familyi={E\bibinitperiod},
           given={Alexandre},
           giveni={A\bibinitperiod}}}%
        {{hash=313d18a818fc7479f0606d72d5b95b81}{%
           family={Sanfourche},
           familyi={S\bibinitperiod},
           given={Martial},
           giveni={M\bibinitperiod}}}%
        {{hash=f6318a99ce304e1f345c1c00415adc8b}{%
           family={Vissiere},
           familyi={V\bibinitperiod},
           given={David},
           giveni={D\bibinitperiod}}}%
        {{hash=b1056fb64ffd8b569084cd51ad8d2c4a}{%
           family={Besnerais},
           familyi={B\bibinitperiod},
           given={Guy},
           giveni={G\bibinitperiod},
           prefix={le},
           prefixi={l\bibinitperiod}}}%
      }
      \strng{namehash}{157b54281ec7f717960ab21c843a9e5a}
      \strng{fullhash}{4dbd5c1f6d33c38c6072503f905278e7}
      \strng{bibnamehash}{4dbd5c1f6d33c38c6072503f905278e7}
      \strng{authorbibnamehash}{4dbd5c1f6d33c38c6072503f905278e7}
      \strng{authornamehash}{157b54281ec7f717960ab21c843a9e5a}
      \strng{authorfullhash}{4dbd5c1f6d33c38c6072503f905278e7}
      \field{extraname}{2}
      \field{sortinit}{2}
      \field{sortinithash}{8b555b3791beccb63322c22f3320aa9a}
      \field{labelnamesource}{author}
      \field{labeltitlesource}{title}
      \field{booktitle}{2017 International Conference on Indoor Positioning and Indoor Navigation (IPIN)}
      \field{title}{An inverse square root filter for robust indoor/outdoor magneto-visual-inertial odometry}
      \field{year}{2017}
      \field{pages}{1\bibrangedash 8}
      \range{pages}{8}
      \verb{doi}
      \verb 10.1109/IPIN.2017.8115888
      \endverb
    \endentry
    \entry{mag_extra_calib}{article}{}
      \name{author}{4}{}{%
        {{hash=2c6be2923f839e167dd99baeb9f09bfd}{%
           family={Gebre-Egziabher},
           familyi={G\bibinithyphendelim E\bibinitperiod},
           given={Demoz},
           giveni={D\bibinitperiod}}}%
        {{hash=73cb398664163c3bc1b57f10bc6140e7}{%
           family={Elkaim},
           familyi={E\bibinitperiod},
           given={Gabriel\bibnamedelima H.},
           giveni={G\bibinitperiod\bibinitdelim H\bibinitperiod}}}%
        {{hash=8c4fd8ce6c465caede7976e20c0f46f1}{%
           family={Powell},
           familyi={P\bibinitperiod},
           given={J.\bibnamedelimi David},
           giveni={J\bibinitperiod\bibinitdelim D\bibinitperiod}}}%
        {{hash=3937c00c0e09ee35964adfa80051463d}{%
           family={Parkinson},
           familyi={P\bibinitperiod},
           given={Bradford\bibnamedelima W.},
           giveni={B\bibinitperiod\bibinitdelim W\bibinitperiod}}}%
      }
      \strng{namehash}{ffcaea274cc51c187bea2b73956a238d}
      \strng{fullhash}{c525679c6df5c6cb584e601686a580aa}
      \strng{bibnamehash}{c525679c6df5c6cb584e601686a580aa}
      \strng{authorbibnamehash}{c525679c6df5c6cb584e601686a580aa}
      \strng{authornamehash}{ffcaea274cc51c187bea2b73956a238d}
      \strng{authorfullhash}{c525679c6df5c6cb584e601686a580aa}
      \field{sortinit}{2}
      \field{sortinithash}{8b555b3791beccb63322c22f3320aa9a}
      \field{labelnamesource}{author}
      \field{labeltitlesource}{title}
      \field{journaltitle}{Journal of Aerospace Engineering}
      \field{number}{2}
      \field{title}{Calibration of Strapdown Magnetometers in Magnetic Field Domain}
      \field{volume}{19}
      \field{year}{2006}
      \field{pages}{87\bibrangedash 102}
      \range{pages}{16}
    \endentry
    \entry{mag_calib}{article}{}
      \name{author}{5}{}{%
        {{hash=e7b3536bb18bb99afc77ea3493c7d196}{%
           family={Vasconcelos},
           familyi={V\bibinitperiod},
           given={J.\bibnamedelimi F.},
           giveni={J\bibinitperiod\bibinitdelim F\bibinitperiod}}}%
        {{hash=c4c9344a1acc2e399c86c29d059f1c99}{%
           family={Elkaim},
           familyi={E\bibinitperiod},
           given={G.},
           giveni={G\bibinitperiod}}}%
        {{hash=738b36d95685dbf9263f8679e563b9bf}{%
           family={Silvestre},
           familyi={S\bibinitperiod},
           given={C.},
           giveni={C\bibinitperiod}}}%
        {{hash=b2e9ebd8d0e6f0a380782b94cddcdd26}{%
           family={Oliveira},
           familyi={O\bibinitperiod},
           given={P.},
           giveni={P\bibinitperiod}}}%
        {{hash=20926fd57ec92a45bbd919ef130268f8}{%
           family={Cardeira},
           familyi={C\bibinitperiod},
           given={B.},
           giveni={B\bibinitperiod}}}%
      }
      \strng{namehash}{e6bc028c67097e3898636c8ccde90719}
      \strng{fullhash}{9fc91cc65a36eba286b75013e775b292}
      \strng{bibnamehash}{9fc91cc65a36eba286b75013e775b292}
      \strng{authorbibnamehash}{9fc91cc65a36eba286b75013e775b292}
      \strng{authornamehash}{e6bc028c67097e3898636c8ccde90719}
      \strng{authorfullhash}{9fc91cc65a36eba286b75013e775b292}
      \field{sortinit}{2}
      \field{sortinithash}{8b555b3791beccb63322c22f3320aa9a}
      \field{labelnamesource}{author}
      \field{labeltitlesource}{title}
      \field{journaltitle}{IEEE Transactions on Aerospace and Electronic Systems}
      \field{number}{2}
      \field{title}{Geometric Approach to Strapdown Magnetometer Calibration in Sensor Frame}
      \field{volume}{47}
      \field{year}{2011}
      \field{pages}{1293\bibrangedash 1306}
      \range{pages}{14}
      \verb{doi}
      \verb 10.1109/TAES.2011.5751259
      \endverb
    \endentry
    \entry{magnetometer_inertial_calibration}{article}{}
      \name{author}{2}{}{%
        {{hash=e54ab5a9a5fa0169ae7f33c4a97f2ea8}{%
           family={Kok},
           familyi={K\bibinitperiod},
           given={Manon},
           giveni={M\bibinitperiod}}}%
        {{hash=a7d921004899fc1272afda1ee990486b}{%
           family={Schön},
           familyi={S\bibinitperiod},
           given={Thomas\bibnamedelima B.},
           giveni={T\bibinitperiod\bibinitdelim B\bibinitperiod}}}%
      }
      \strng{namehash}{2b13cd820653164d2d09e99dbbf63e96}
      \strng{fullhash}{2b13cd820653164d2d09e99dbbf63e96}
      \strng{bibnamehash}{2b13cd820653164d2d09e99dbbf63e96}
      \strng{authorbibnamehash}{2b13cd820653164d2d09e99dbbf63e96}
      \strng{authornamehash}{2b13cd820653164d2d09e99dbbf63e96}
      \strng{authorfullhash}{2b13cd820653164d2d09e99dbbf63e96}
      \field{sortinit}{2}
      \field{sortinithash}{8b555b3791beccb63322c22f3320aa9a}
      \field{labelnamesource}{author}
      \field{labeltitlesource}{title}
      \field{journaltitle}{IEEE Sensors Journal}
      \field{number}{14}
      \field{title}{Magnetometer Calibration Using Inertial Sensors}
      \field{volume}{16}
      \field{year}{2016}
      \field{pages}{5679\bibrangedash 5689}
      \range{pages}{11}
      \verb{doi}
      \verb 10.1109/JSEN.2016.2569160
      \endverb
    \endentry
    \entry{solin_gaussian}{article}{}
      \name{author}{5}{}{%
        {{hash=4592dc3c98e842594b9416db6b6470ae}{%
           family={Solin},
           familyi={S\bibinitperiod},
           given={Arno},
           giveni={A\bibinitperiod}}}%
        {{hash=e54ab5a9a5fa0169ae7f33c4a97f2ea8}{%
           family={Kok},
           familyi={K\bibinitperiod},
           given={Manon},
           giveni={M\bibinitperiod}}}%
        {{hash=f21f2855787c39337ef57063882d1a29}{%
           family={Wahlström},
           familyi={W\bibinitperiod},
           given={Niklas},
           giveni={N\bibinitperiod}}}%
        {{hash=a7d921004899fc1272afda1ee990486b}{%
           family={Schön},
           familyi={S\bibinitperiod},
           given={Thomas\bibnamedelima B.},
           giveni={T\bibinitperiod\bibinitdelim B\bibinitperiod}}}%
        {{hash=89f4b98fe43f3c9f7ffdc081904d4781}{%
           family={Särkkä},
           familyi={S\bibinitperiod},
           given={Simo},
           giveni={S\bibinitperiod}}}%
      }
      \strng{namehash}{ecbb0100ca6d058cc81ce2592acb6ffa}
      \strng{fullhash}{5d0b3c7916a157e7e24bc023c7560be4}
      \strng{bibnamehash}{5d0b3c7916a157e7e24bc023c7560be4}
      \strng{authorbibnamehash}{5d0b3c7916a157e7e24bc023c7560be4}
      \strng{authornamehash}{ecbb0100ca6d058cc81ce2592acb6ffa}
      \strng{authorfullhash}{5d0b3c7916a157e7e24bc023c7560be4}
      \field{sortinit}{2}
      \field{sortinithash}{8b555b3791beccb63322c22f3320aa9a}
      \field{labelnamesource}{author}
      \field{labeltitlesource}{title}
      \field{journaltitle}{IEEE Transactions on Robotics}
      \field{number}{4}
      \field{title}{Modeling and Interpolation of the Ambient Magnetic Field by Gaussian Processes}
      \field{volume}{34}
      \field{year}{2018}
      \field{pages}{1112\bibrangedash 1127}
      \range{pages}{16}
      \verb{doi}
      \verb 10.1109/TRO.2018.2830326
      \endverb
    \endentry
    \entry{quatenrion_sola}{misc}{}
      \name{author}{1}{}{%
        {{hash=e082ff94720eff98805e9e08ca921bcc}{%
           family={Solà},
           familyi={S\bibinitperiod},
           given={Joan},
           giveni={J\bibinitperiod}}}%
      }
      \strng{namehash}{e082ff94720eff98805e9e08ca921bcc}
      \strng{fullhash}{e082ff94720eff98805e9e08ca921bcc}
      \strng{bibnamehash}{e082ff94720eff98805e9e08ca921bcc}
      \strng{authorbibnamehash}{e082ff94720eff98805e9e08ca921bcc}
      \strng{authornamehash}{e082ff94720eff98805e9e08ca921bcc}
      \strng{authorfullhash}{e082ff94720eff98805e9e08ca921bcc}
      \field{sortinit}{2}
      \field{sortinithash}{8b555b3791beccb63322c22f3320aa9a}
      \field{labelnamesource}{author}
      \field{labeltitlesource}{title}
      \field{title}{Quaternion kinematics for the error-state Kalman filter}
      \field{year}{2017}
      \verb{eprint}
      \verb arXiv:1711.02508
      \endverb
    \endentry
    \entry{kalibr}{inproceedings}{}
      \name{author}{3}{}{%
        {{hash=9f26483b21e967b59019c49805811483}{%
           family={Furgale},
           familyi={F\bibinitperiod},
           given={Paul},
           giveni={P\bibinitperiod}}}%
        {{hash=ebb4c8c255c177b810c070b09d29426f}{%
           family={Rehder},
           familyi={R\bibinitperiod},
           given={Joern},
           giveni={J\bibinitperiod}}}%
        {{hash=6ee8958bbf32527489012d8ca7c95ee3}{%
           family={Siegwart},
           familyi={S\bibinitperiod},
           given={Roland},
           giveni={R\bibinitperiod}}}%
      }
      \strng{namehash}{f65a74c390ff5e9ae01fdc843d9b1fbf}
      \strng{fullhash}{54c4383ce461e1a8f2d94e052ba81430}
      \strng{bibnamehash}{54c4383ce461e1a8f2d94e052ba81430}
      \strng{authorbibnamehash}{54c4383ce461e1a8f2d94e052ba81430}
      \strng{authornamehash}{f65a74c390ff5e9ae01fdc843d9b1fbf}
      \strng{authorfullhash}{54c4383ce461e1a8f2d94e052ba81430}
      \field{sortinit}{3}
      \field{sortinithash}{ad6fe7482ffbd7b9f99c9e8b5dccd3d7}
      \field{labelnamesource}{author}
      \field{labeltitlesource}{title}
      \field{booktitle}{2013 IEEE/RSJ International Conference on Intelligent Robots and Systems}
      \field{title}{Unified temporal and spatial calibration for multi-sensor systems}
      \field{year}{2013}
      \field{pages}{1280\bibrangedash 1286}
      \range{pages}{7}
      \verb{doi}
      \verb 10.1109/IROS.2013.6696514
      \endverb
    \endentry
    \entry{ieee_imu_standard}{article}{}
      \field{sortinit}{3}
      \field{sortinithash}{ad6fe7482ffbd7b9f99c9e8b5dccd3d7}
      \field{labeltitlesource}{title}
      \field{journaltitle}{IEEE Std 952-2020 (Revision of IEEE Std 952-1997)}
      \field{title}{{IEEE} Standard for Specifying and Testing Single-Axis Interferometric Fiber Optic Gyros}
      \field{year}{2021}
      \field{pages}{1\bibrangedash 93}
      \range{pages}{93}
      \verb{doi}
      \verb 10.1109/IEEESTD.2021.9353434
      \endverb
    \endentry
    \entry{mag_ellipsoid}{article}{}
      \name{author}{2}{}{%
        {{hash=c84333257e5fef7e25ef4f7a4111077c}{%
           family={Wu},
           familyi={W\bibinitperiod},
           given={Y.},
           giveni={Y\bibinitperiod}}}%
        {{hash=912a43b3957453af94e052366373879e}{%
           family={Shi},
           familyi={S\bibinitperiod},
           given={W.},
           giveni={W\bibinitperiod}}}%
      }
      \strng{namehash}{43a8d78af0598ad58f4f739c840af115}
      \strng{fullhash}{43a8d78af0598ad58f4f739c840af115}
      \strng{bibnamehash}{43a8d78af0598ad58f4f739c840af115}
      \strng{authorbibnamehash}{43a8d78af0598ad58f4f739c840af115}
      \strng{authornamehash}{43a8d78af0598ad58f4f739c840af115}
      \strng{authorfullhash}{43a8d78af0598ad58f4f739c840af115}
      \field{sortinit}{4}
      \field{sortinithash}{9381316451d1b9788675a07e972a12a7}
      \field{labelnamesource}{author}
      \field{labeltitlesource}{title}
      \field{journaltitle}{IEEE Sensors Journal}
      \field{number}{11}
      \field{title}{On Calibration of Three-Axis Magnetometer}
      \field{volume}{15}
      \field{year}{2015}
      \field{pages}{6424\bibrangedash 6431}
      \range{pages}{8}
      \verb{doi}
      \verb 10.1109/JSEN.2015.2459767
      \endverb
    \endentry
    \entry{umeyama}{article}{}
      \name{author}{1}{}{%
        {{hash=2dc7cd18759a5a030ee67028660a94ff}{%
           family={Umeyama},
           familyi={U\bibinitperiod},
           given={S.},
           giveni={S\bibinitperiod}}}%
      }
      \strng{namehash}{2dc7cd18759a5a030ee67028660a94ff}
      \strng{fullhash}{2dc7cd18759a5a030ee67028660a94ff}
      \strng{bibnamehash}{2dc7cd18759a5a030ee67028660a94ff}
      \strng{authorbibnamehash}{2dc7cd18759a5a030ee67028660a94ff}
      \strng{authornamehash}{2dc7cd18759a5a030ee67028660a94ff}
      \strng{authorfullhash}{2dc7cd18759a5a030ee67028660a94ff}
      \field{sortinit}{4}
      \field{sortinithash}{9381316451d1b9788675a07e972a12a7}
      \field{labelnamesource}{author}
      \field{labeltitlesource}{title}
      \field{journaltitle}{IEEE Transactions on Pattern Analysis and Machine Intelligence}
      \field{number}{4}
      \field{title}{Least-squares estimation of transformation parameters between two point patterns}
      \field{volume}{13}
      \field{year}{1991}
      \field{pages}{376\bibrangedash 380}
      \range{pages}{5}
      \verb{doi}
      \verb 10.1109/34.88573
      \endverb
    \endentry
    \entry{absolute_traj_error}{inproceedings}{}
      \name{author}{5}{}{%
        {{hash=9e291584a3b29b27c88ecfe0a566274c}{%
           family={Sturm},
           familyi={S\bibinitperiod},
           given={Jürgen},
           giveni={J\bibinitperiod}}}%
        {{hash=9d6110950a50d69179aefe99b3abc81a}{%
           family={Engelhard},
           familyi={E\bibinitperiod},
           given={Nikolas},
           giveni={N\bibinitperiod}}}%
        {{hash=4858aa954321596e8c1c81daac0271dd}{%
           family={Endres},
           familyi={E\bibinitperiod},
           given={Felix},
           giveni={F\bibinitperiod}}}%
        {{hash=98f4ab1bc2ac191d8f8a0651315ce9c0}{%
           family={Burgard},
           familyi={B\bibinitperiod},
           given={Wolfram},
           giveni={W\bibinitperiod}}}%
        {{hash=1bd2b6b6ca2fc15a90f164070b626131}{%
           family={Cremers},
           familyi={C\bibinitperiod},
           given={Daniel},
           giveni={D\bibinitperiod}}}%
      }
      \strng{namehash}{f2452d6a1e4ac6399e084b1a065ee7da}
      \strng{fullhash}{fda811f69253ca27ebab9abbbf7b757f}
      \strng{bibnamehash}{fda811f69253ca27ebab9abbbf7b757f}
      \strng{authorbibnamehash}{fda811f69253ca27ebab9abbbf7b757f}
      \strng{authornamehash}{f2452d6a1e4ac6399e084b1a065ee7da}
      \strng{authorfullhash}{fda811f69253ca27ebab9abbbf7b757f}
      \field{sortinit}{5}
      \field{sortinithash}{20e9b4b0b173788c5dace24730f47d8c}
      \field{labelnamesource}{author}
      \field{labeltitlesource}{title}
      \field{booktitle}{2012 IEEE/RSJ International Conference on Intelligent Robots and Systems}
      \field{title}{A benchmark for the evaluation of RGB-D SLAM systems}
      \field{year}{2012}
      \field{pages}{573\bibrangedash 580}
      \range{pages}{8}
      \verb{doi}
      \verb 10.1109/IROS.2012.6385773
      \endverb
    \endentry
    \entry{Zhang18iros}{inproceedings}{}
      \name{author}{2}{}{%
        {{hash=33393646bb4bfe56c3bb0a2ec46223dc}{%
           family={Zhang},
           familyi={Z\bibinitperiod},
           given={Zichao},
           giveni={Z\bibinitperiod}}}%
        {{hash=dac21ab4ede215439fcc6b051be53a11}{%
           family={Scaramuzza},
           familyi={S\bibinitperiod},
           given={Davide},
           giveni={D\bibinitperiod}}}%
      }
      \strng{namehash}{e8849a46e9d692dc9c0a96307e1297f8}
      \strng{fullhash}{e8849a46e9d692dc9c0a96307e1297f8}
      \strng{bibnamehash}{e8849a46e9d692dc9c0a96307e1297f8}
      \strng{authorbibnamehash}{e8849a46e9d692dc9c0a96307e1297f8}
      \strng{authornamehash}{e8849a46e9d692dc9c0a96307e1297f8}
      \strng{authorfullhash}{e8849a46e9d692dc9c0a96307e1297f8}
      \field{sortinit}{5}
      \field{sortinithash}{20e9b4b0b173788c5dace24730f47d8c}
      \field{labelnamesource}{author}
      \field{labeltitlesource}{title}
      \field{booktitle}{IEEE/RSJ Int. Conf. Intell. Robot. Syst. (IROS)}
      \field{title}{A Tutorial on Quantitative Trajectory Evaluation for Visual(-Inertial) Odometry}
      \field{year}{2018}
    \endentry
    \entry{bluerov}{misc}{}
      \name{author}{1}{}{%
        {{hash=9528f85f2c807e7a922065c08d7c5c82}{%
           family={Robotics},
           familyi={R\bibinitperiod},
           given={Blue},
           giveni={B\bibinitperiod}}}%
      }
      \strng{namehash}{9528f85f2c807e7a922065c08d7c5c82}
      \strng{fullhash}{9528f85f2c807e7a922065c08d7c5c82}
      \strng{bibnamehash}{9528f85f2c807e7a922065c08d7c5c82}
      \strng{authorbibnamehash}{9528f85f2c807e7a922065c08d7c5c82}
      \strng{authornamehash}{9528f85f2c807e7a922065c08d7c5c82}
      \strng{authorfullhash}{9528f85f2c807e7a922065c08d7c5c82}
      \field{sortinit}{5}
      \field{sortinithash}{20e9b4b0b173788c5dace24730f47d8c}
      \field{labelnamesource}{author}
      \field{labeltitlesource}{title}
      \field{howpublished}{\url{https://bluerobotics.com/store/rov/bluerov2/}}
      \field{title}{{BlueROV2}}
      \field{year}{2023}
    \endentry
    \entry{DudekIROS2005}{inproceedings}{}
      \name{author}{15}{}{%
        {{hash=13d3850176ca9bbfb14e68325cc0a9fc}{%
           family={Dudek},
           familyi={D\bibinitperiod},
           given={Gregory},
           giveni={G\bibinitperiod}}}%
        {{hash=66811e93f6838b37e0fd6cb2aad9fc05}{%
           family={Jenkin},
           familyi={J\bibinitperiod},
           given={Michael},
           giveni={M\bibinitperiod}}}%
        {{hash=fa861d5a90d8f50b1ad17d5930a1ffa7}{%
           family={Prahacs},
           familyi={P\bibinitperiod},
           given={Chriss},
           giveni={C\bibinitperiod}}}%
        {{hash=9302a8419fe15a03aa454bc07373c396}{%
           family={Hogue},
           familyi={H\bibinitperiod},
           given={Andrew},
           giveni={A\bibinitperiod}}}%
        {{hash=57f3d65edf65578d5be03190d6170c56}{%
           family={Sattar},
           familyi={S\bibinitperiod},
           given={Junaed},
           giveni={J\bibinitperiod}}}%
        {{hash=f6fe96d5fbd99848c75f7e75975bf475}{%
           family={Giguere},
           familyi={G\bibinitperiod},
           given={Philippe},
           giveni={P\bibinitperiod}}}%
        {{hash=e4e245194497f42ce8249ec6c12eaf0a}{%
           family={German},
           familyi={G\bibinitperiod},
           given={Andrew},
           giveni={A\bibinitperiod}}}%
        {{hash=7135a2c35bf03306d6346621a8c28a10}{%
           family={Liu},
           familyi={L\bibinitperiod},
           given={Hongyu},
           giveni={H\bibinitperiod}}}%
        {{hash=d2ebb4399063c22c3584d0856b610859}{%
           family={Saunderson},
           familyi={S\bibinitperiod},
           given={Shane},
           giveni={S\bibinitperiod}}}%
        {{hash=a1792e6e4b9f0204391b07a22e9fedc5}{%
           family={Ripsman},
           familyi={R\bibinitperiod},
           given={Arlene},
           giveni={A\bibinitperiod}}}%
        {{hash=90a891c504fa5facc17214121b4c72dd}{%
           family={Simhon},
           familyi={S\bibinitperiod},
           given={Saul},
           giveni={S\bibinitperiod}}}%
        {{hash=ef2cffedfb3e90bb9020e325bb215021}{%
           family={Torres-Mendez},
           familyi={T\bibinithyphendelim M\bibinitperiod},
           given={Luz\bibnamedelima Abril},
           giveni={L\bibinitperiod\bibinitdelim A\bibinitperiod}}}%
        {{hash=439644dae7024652508e5642aa405cf9}{%
           family={Milios},
           familyi={M\bibinitperiod},
           given={Evangelos},
           giveni={E\bibinitperiod}}}%
        {{hash=0a0059ced34350551e0338dfbbbee893}{%
           family={Zhang},
           familyi={Z\bibinitperiod},
           given={Pifu},
           giveni={P\bibinitperiod}}}%
        {{hash=b47c2dc8dc3c61fadae36fa40955b725}{%
           family={Rekleitis},
           familyi={R\bibinitperiod},
           given={Ioannis},
           giveni={I\bibinitperiod}}}%
      }
      \list{location}{1}{%
        {Edmonton AB, Canada}%
      }
      \strng{namehash}{9c7bbb3c0c8aa47ce2f75b4ab3c2e6af}
      \strng{fullhash}{834f62e24f7a312f0e93946170ecac60}
      \strng{bibnamehash}{9d1473ff56ef918004bb618692f40e77}
      \strng{authorbibnamehash}{9d1473ff56ef918004bb618692f40e77}
      \strng{authornamehash}{9c7bbb3c0c8aa47ce2f75b4ab3c2e6af}
      \strng{authorfullhash}{834f62e24f7a312f0e93946170ecac60}
      \field{sortinit}{5}
      \field{sortinithash}{20e9b4b0b173788c5dace24730f47d8c}
      \field{labelnamesource}{author}
      \field{labeltitlesource}{title}
      \field{abstract}{We describe recent results obtained with AQUA, a mobile robot capable of swimming, walking and amphibious operation. Designed to rely primarily on visual sensors, the AQUA robot uses vision to navigate underwater using servobased guidance, and also to obtain high-resolution range scans of its local environment. This paper describes some of the pragmatic and logistic obstacles encountered, and provides an overview of some of the basic capabilities of the vehicle and its associated sensors. Moreover, this paper presents the first ever amphibious transition from walking to swimming.}
      \field{booktitle}{IEEE/RSJ International Conference on Intelligent Robots and Systems (IROS)}
      \field{label}{C22}
      \field{month}{8}
      \field{title}{A Visually Guided Swimming Robot}
      \field{year}{2005}
      \field{pages}{1749\bibrangedash 1754}
      \range{pages}{6}
      \verb{file}
      \verb iros05-1053.pdf
      \endverb
    \endentry
  \enddatalist
\endrefsection

\begin{minipage}{0.90\textwidth}\ \\[12pt]
    \vspace{3in}
    \begin{center}
        This paper has been accepted for publication in \textit{IEEE Conference on Robotics and Automation 2024}.
    \end{center}

    \vspace{1in}
    ©2024 IEEE. Personal use of this material is permitted. Permission from IEEE must be obtained for all other uses, in any current or future media, including reprinting/republishing this material for advertising or promotional purposes, creating new collective works, for resale or redistribution to servers or lists, or reuse of any copyrighted component of this work in other works.
\end{minipage}

\newpage

\maketitle
\begin{abstract}
    This paper presents an extension to visual inertial odometry (VIO) by introducing tightly-coupled fusion of magnetometer measurements. A sliding window of keyframes is optimized by minimizing re-projection errors, relative inertial errors, and relative magnetometer orientation errors. The results of IMU orientation propagation are used to efficiently transform magnetometer measurements between frames producing relative orientation constraints between consecutive frames. The soft and hard iron effects are calibrated using an ellipsoid fitting algorithm. The introduction of magnetometer data results in significant reductions in the orientation error and also in recovery of the true yaw orientation with respect to the magnetic north. The proposed framework operates in all environments with slow-varying magnetic fields, mainly outdoors and underwater. We have focused our work on the underwater domain, especially in underwater caves, as the narrow passage and turbulent flow make it difficult to perform loop closures and reset the localization drift. The underwater caves present challenges to VIO due to the absence of ambient light and the confined nature of the environment, while also being a crucial source of fresh water and providing valuable historical records. Experimental results from underwater caves demonstrate the improvements in accuracy and robustness introduced by the proposed VIO extension.
\end{abstract}

\section{Introduction}

\begin{figure}[t]
    \centering
    {\includegraphics[width=0.9\columnwidth]{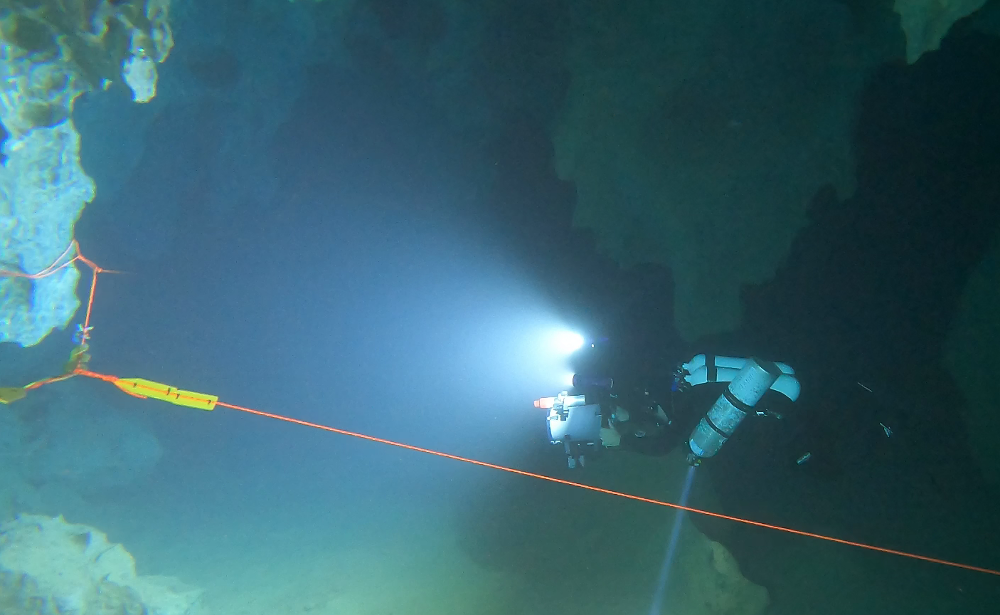}}
    \caption{Sensor suite deployed inside the Cueva del Agua, Spain.}\vspace{-0.3in}
    \label{fig:beauty}
\end{figure}

Magnetic measurements are an often neglected source of information mainly because of their sensitivity to ambient noise; however, there are several situations in which they can provide useful information with minimal cost and low computational overhead.  In this work we are targeting the underwater domain with the use of a sensor suite~\cite{RahmanOceans2018}. The used sensor suite, comprised of a stereo camera, a 9-axis IMU, a water depth sensor, and a pencil beam mechanically scanning sonar, can be deployed by a human to collect data inside an underwater cave; see ~\fig{~\ref{fig:beauty}}. We propose a tightly coupled optimization-based fusion of visual, inertial, and magnetometer information. Magnetometer measurements are added as new factors to the keyframe-based sliding window optimization graph proposed in ~\cite{okvis}. We leverage the IMU preintegration algorithm from~\cite{qin2017vins_mono, forster2016manifold} to efficiently compute magnetometer residuals for all measurements. Since the IMU preintegration terms are already defined in the relative inertial error, the additional computational cost is relatively small. The magnetometer data are sensitive to the local magnetic field and require an explicit calibration procedure. As such, the magnetometer measurements are calibrated to account for soft and hard iron effects using an ellipsoid fitting algorithm. Experiments have shown that after calibration, the local magnetic noise is relatively low throughout the trajectory. Furthermore, magnetic data present an absolute measurement based on the magnetic field of Earth and while noisy they are consistent over long trajectories. Our target application is the mapping of underwater caves.

Mapping underwater cave systems is extremely important for environmental protection, fresh water management~\cite{Florida2010}, and resource utilization~\cite{ZexuanReports2016}. Moreover, caves provide valuable historical evidence as they present an undisturbed time capsule~\cite{Abbott2014,macdonald2020paleoindian,rissolo2015novel,de2015ancient}, and information about geological processes~\cite{KresicMikszewski2013}. Diver centered mapping is a dangerous, labor\hyp intensive, tedious, and slow task. \invis{Streamlining the process by having a sensor do the majority of the mapping tasks will reduce the cognitive load enabling the divers to concentrate on the navigation decisions, potentially reducing the often lethal risks (more than $350$ deaths since $1969$ in USA alone~\cite{potts2016thirty,buzzacott2017recovery}). Furthermore, underwater cave environments present additional challenges compared to the standard underwater domain, affecting both divers and deployed sensors.} In addition to limited visibility, color absorption, hazing, and lightning variations, there is no ambient light and the environment is often cluttered and fragile, making navigation difficult, with inadvertent motion maneuvers often reducing visibility to zero. Furthermore, the confined environment prevents resurfacing in case of a problem or emergency. During operations inside underwater caves, there are two constraining factors. First, the passages are often narrow, and the view on the way in is completely different from the view on the way out, making maneuvering difficult. Second, there is significant water flow that makes staying in one spot to collect data in 360 degrees very challenging. These conditions contribute to reducing the number of loop closures that are feasible.

Moreover, underwater cave systems provide an exciting opportunity to use magnetometer measurements as they are devoid of significant magnetic disturbances due to ferromagnetic materials. The introduction of magnetic field measurements in the state estimation process, proposed in this paper, leads to two significant contributions. First of all, absolute orientation measurements are introduced to orient the produced trajectory with respect to the magnetic north. This makes the produced trajectories compatible with the existing man\hyp made maps of the caves. Second, the introduction of magnetometer data constrains the produced trajectories along the yaw direction, producing much more consistent results, eliminating the orientation drift that plagued earlier deployments~\cite{JoshiAUV2022}.

To the best of our knowledge, our work is the first to use IMU preintegration to introduce high frequency magnetometer measurements in the optimization framework. The proposed framework has been tested in a variety of underwater caves in Florida and Mexico. Qualitative results demonstrate the alignment of the produced trajectories to the existing maps of the caves, and also a significant reduction in orientation error (drift). The produced trajectories resulted in significantly reduced error, compared to the baseline trajectories obtained using COLMAP~\cite{colmap_sfm} (a bundle adjustment, global optimization framework).

\section{Related Work}
Vision\hyp based state estimation has proven to be extremely challenging due to the varying lightning conditions, scattering and haziness from floating particulates, and color variations due to light absorption in the water. Several visual and visual-inertial state estimation packages have exhibited severe failures~\cite{QuattriniLiISERVO2016,JoshiIROS2019}. Rahman \etal~\cite{RahmanICRA2018} proposed an extension of OKVIS~\cite{okvis} incorporating a pencil beam mechanically scanning sonar, a water depth sensor, and loop closure capabilities~\cite{RahmanIROS2019a,RahmanIJRR2022}. The framework was then adapted for inexpensive action cameras, GoPro 9,~\cite{JoshiICRA2022} producing superior performance. More recent packages such as ORB-SLAM3~\cite{orbslam3}, OpenVINS~\cite{geneva2020openvins},  \etc have demonstrated better performance, but there is no in-depth analysis of their accuracy and robustness.

Magnetometers have traditionally been fused with acceleration and angular velocity measurements to estimate the vehicle's attitude; such systems are termed attitude and heading reference systems (AHRS) ~\cite{madwick_filter, complimentary_filter, zhao_imu_mag_ultrasonic}. Some recent work on the fusion of magnetic field measurements with a visual sensor has gained traction. Wang \etal ~\cite{wang_vimo} performed visual, inertial, and magnetometer fusion first by initializing the VIO in the reference frame of Earth and then computing the error between magnetometer measurement and Earth's magnetic field. The authors included only magnetometer measurements near the keyframe, skipping most of the high-frequency magnetometer measurements. Siebler \etal proposed a particle filter method to use magnetic field distortions for magnetometer calibration and localization ~\cite{Siebler}; without focusing on continuous state estimation. The closest approach to ours used inter-frame preintegrated magnetometer measurements~\cite{caruso_mimu, caruso_mimu_kf}. However, it requires an array of magnetometers measuring magnetic field and it's gradient. Our work is focused on a single 9-axis IMU without requiring additional hardware enhancements.

The calibration of soft and hard iron offsets for magnetometer is very important as they can introduce huge errors in the estimation process. In ~\cite{mag_extra_calib} an iterative batch least squares estimation was introduced to calibrate a full three-axis magnetometer. Vasconcelos \etal used maximum likelihood estimation to find the optimal calibration parameters that best fit the reading of the onboard sensors to calibrate the 3-axis strapless magnetometer ~\cite{mag_calib}. Kok \etal ~\cite{magnetometer_inertial_calibration} proposed maximal likelihood calibration using orientation estimation from inertial sensors. Taking things to the next step, Solin \etal created a map of the magnetic field in an indoor environment by imposing a Gaussian process (GP) prior to the latent scalar potential of the magnetic field~\cite{solin_gaussian}.

\section{Visual-Inertial-Magnetometer Fusion}
\label{sec:theory}
Visual-Inertial Odometry aims to estimate the pose of the moving camera-imu system by fusing inertial information with images. In this work, we extend visual-inertial odometry to fuse magnetometer measurements in a tightly coupled fashion using IMU preintegration.

\subsection{Problem Formulation}
\label{sec:problem_formulation}
\begin{figure}
    \centering
    \vspace{0.1in}
    \includegraphics[width=0.7\columnwidth]{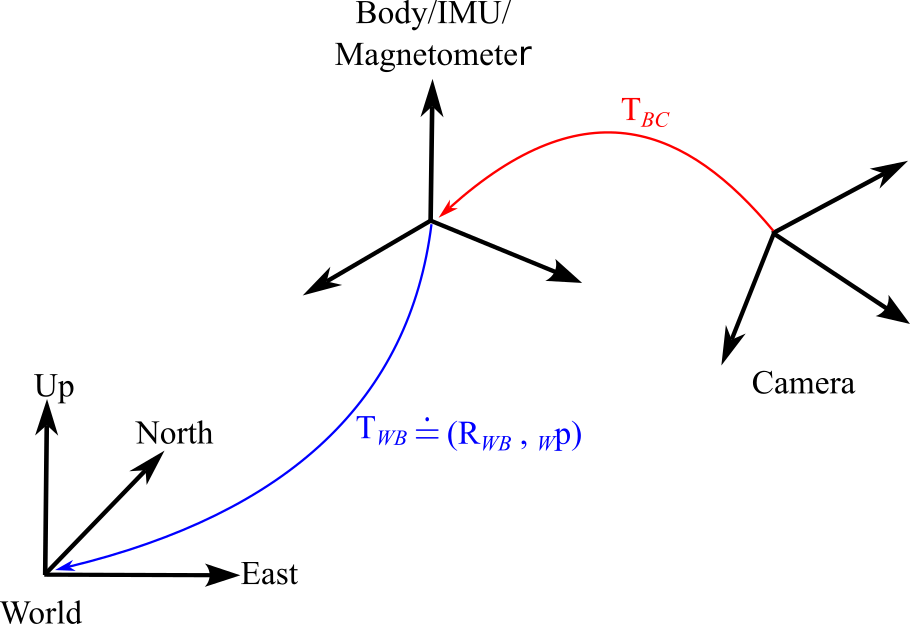}
    \caption{The body pose $\text{T}_\text{WB} \doteq (\text{R}_\text{WB}, {}_\text{W}\textbf{p}$) which coincides with 9-axis IMU is tracked w.r.t the world frame W. The camera pose in body frame $\text{T}_{BC}$ is fixed and known from prior calibration. The world frame W coincides with East, North and Up direction in Earth coordinates.}\vspace{-0.3in}
    \label{fig:coordinate_frames}
\end{figure}

We consider the visual inertial magnetometer odometry problem where we want to track the \textit{state} of a sensing system (a handheld underwater sensor suite) equipped with an IMU, a stereo/monocular camera, and a magnetometer. The IMU frame coincides with the body frame "B" we want to track, and the transformation between the camera and the IMU ($\text{T}_\text{BC}$) is known by calibrating the extrinsic parameters and remains constant throughout the experiment; see ~\fig{\ref{fig:coordinate_frames}} . In this work, we use a 9-axis IMU that includes an accelerometer, a gyroscope, and a magnetometer. As such, we assume that the magnetometer and IMU (gyroscope and accelerometer) axes are aligned and that there is no significant axis misalignment. When using stereo cameras, we assume that they are rigidly attached with known extrinsic parameters between the cameras and between the cameras and the IMU.

Nonlinear keyframe-based sliding window optimization is performed to estimate body poses and 3D landmark positions by minimizing the reprojection error of landmarks seen in the camera frame. Thus, we need to estimate the variables $\pazocal{X} = \{\pazocal{X}_{\text{B}}, \pazocal{L}\}$ where $\pazocal{L}$ represents 3D landmark positions visible in the sliding window and $\pazocal{X}_{\text{B}} = [\mathbf{x}_1, ..., \mathbf{x}_K]$ holds the system states at camera times 1 to $K$ with $K$ being the total number of keyframes in the sliding window. The system state $\mathbf{x}_{k}$ at time $t_k$ holds position ${}_{\text{W}}\mathbf{p}^k$ and orientation $\mathbf{q}_{\text{WB}}^k$ in the world frame, velocity in the inertial frame ${}_\text{W}\mathbf{v}^k$, as well as biases of the gyroscope $\mathbf{b}_g^k$ and accelerometer $\mathbf{b}_a^k$. The system state can be written as
\begin{equation}
    \mathbf{x}_{k} = [{}_\text{W}\mathbf{p}^k, \mathbf{q}_{\text{WB}}^k, {}_\text{W}\mathbf{v}^k, \mathbf{b}_g^k, \mathbf{b}_a^k]
    \label{eq:state}
\end{equation}
where $\mathbf{q}_\text{WB}$ is the quaternion representation of the orientation $\text{R}_{\text{WB}}$.

\subsection{Fusing visual, inertial and magnetometer measurements}
\label{sec:vim-fusion}
The keyframe-based visual-inertial SLAM is formulated as joint nonlinear optimization that maximizes the posterior probability of the system state $\pazocal{X}$. Using the problem formulation proposed in ~\cite{okvis, RahmanIJRR2022}, the following cost function is minimized
\begin{equation}
    J_{VI}(\pazocal{X}) = \sum_{k=1}^{K}\sum_{j\in \pazocal{J}_k} \lVert \mathbf{e_r}^{j,k} \rVert^2_{\mathbf{W_r}^{j,k}} + \sum_{k=1}^{K} \lVert \mathbf{e_i}^{k} \rVert^2_{\mathbf{W_i}^{k}} + \lVert \mathbf{e_p}\rVert^2
    \label{eq:vio_costfunction}
\end{equation}
where $k$ denotes the camera frame index and $j$ denotes the landmark index. The cost function contains the reprojection $\mathbf{e_r}$, inertial $\mathbf{e_i}$, and marginalization $\mathbf{e_p}$ residuals weighted by their respective information matrices.

The reprojection error for the landmark ${}_W \mathbf{l}_j \in \pazocal{J}_k$ is calculated as $\mathbf{e_r}^{i,j} = h({}_W \mathbf{l}_j) - \mathbf{z}^{j,k}$ where $\pazocal{J}_k$ is the set of all the landmarks visible in the keyframe $k$. Here, $h(.)$ denotes the camera projection model and $\mathbf{z}^{j,k}$ the measurements in the image coordinates. For more details, please refer to ~\cite{okvis}. The inertial residuals $\mathbf{e_i}$ are obtained using the IMU preintegration theory proposed in ~\cite{qin2017vins_mono, quatenrion_sola} which is detailed in ~\sect{~\ref{sec:imu_preintegration}}. We employ a marginalization strategy similar to ~\cite{okvis} to obtain the marginalization prior error term, $\mathbf{e_p}$. Whenever a new frame is inserted in the optimization window, the marginalization operation is classified into two cases. If the oldest frame in the sliding window is not a keyframe, it is marginalized together with the oldest speed and bias states, and all its landmark measurements are dropped to maintain sparsity. In the case where the oldest frame is a keyframe, only the landmarks that are visible in that frame but not in the most recent keyframe are marginalized out.

We introduce the magnetometer residuals based on IMU preintegration into ~\eq{~\ref{eq:vio_costfunction}} to obtain the cost function used in this work as
\begin{equation}
    J(\pazocal{X}) = J_{VI}(\pazocal{X}) + \sum_{k=1}^{K}\sum_{j\in \pazocal{M}_k} \lVert \mathbf{e_m}^{j,k} \rVert^2_{\mathbf{W_m}^{k}} \label{eq:costfunction}
\end{equation}
where $\pazocal{M}_k$ denotes all magnetometer measurements attached to the system state $\mathbf{x}_k$ by the error term. In the remainder of ~\sect{~\ref{sec:theory}}, we use the output of the IMU preintegration algorithm to derive the magnetometer error term $\mathbf{e_m}^{j,k}$ and the residual weights $\mathbf{W_m}^k$.

\subsection{IMU Preintegration}
\label{sec:imu_preintegration}

In this section, we review the IMU preintegration formulation, which in turn will be used to derive the magnetometer residuals in ~\sect{~\ref{sec:magnetometer_residuals}}. The IMU preintegration formulation is based on ~\cite{qin2017vins_mono} inspired by continuous-time quaternion kinematics from ~\cite{quatenrion_sola}  and uses IMU bias manipulation according to ~\cite{forster2016manifold}.

The accelerometer and gyroscope measurements in body frame B at time $t$ are affected by the additive white noise $\boldsymbol{\eta}$ and slowing varying bias $\mathbf{b}$
\begin{equation}
    \label{eq:imu_measurements}
    \begin{aligned}
        {}_\text{B}\tilde{\boldsymbol{w}}(t) & =  {}_\text{B}\boldsymbol{w}(t) + \mathbf{b}_g(t) + \boldsymbol{\eta}_g                                                  \\
        {}_\text{B}\tilde{\mathbf{a}}(t)     & =  \text{R}_\text{WB}^\text{T}({}_\text{W}\mathbf{a}(t) - {}_\text{W}\mathbf{g}) + \mathbf{b}_a(t) + \boldsymbol{\eta}_a
    \end{aligned}
\end{equation}

The accelerometer and gyroscope noise is modeled as additive Gaussian noise with $\boldsymbol{\eta}_a \sim \pazocal{N}(\mathbf{0}, \sigma_a^2 \cdot \mathbf{I}) $ and $\boldsymbol{\eta}_w \sim \pazocal{N}(\mathbf{0}, \sigma_w^2 \cdot \mathbf{I}) $, where $\mathbf{I}$ being the identity matrix.  IMU biases are modeled as a slowly varying random walk with $\dot{\mathbf{b}}_{w_t} = \boldsymbol{\eta}_{b_w}$ and $\dot{\mathbf{b}}_{a_t} = \boldsymbol{\eta}_{b_a}$ where $\boldsymbol{\eta}_{b_w} \sim \pazocal{N}(\mathbf{0}, \sigma_{b_w}^2 \cdot \mathbf{I}) $ and $\boldsymbol{\eta}_{b_a} \sim \pazocal{N}(\mathbf{0}, \sigma_{b_a}^2 \cdot \mathbf{I}) $. Most modern 9-axis IMUs provide high-frequency synchronized accelerometer, gyroscope, and magnetometer measurements, as shown in Fig.~\ref{fig:visual_inertial_magnetometer_timing}. As such, ${}_\text{W}\mathbf{p}^i, \mathbf{q}_\text{WB}^i$ and ${}_\text{W}\mathbf{v}^i$ can be propagated in the time interval $[t_k, t_{k+1}]$ using accelerometer and gyroscope measurements. Although in practice IMU measurements may not perfectly synchronize with image timestamps, the sensor setup is calibrated using Kalibr~\cite{kalibr} beforehand.  Propagation in the world frame requires knowledge of the initial state of the system at time $t_k$. Whenever the system state changes during optimization, repropagation is needed. Thus, the IMU preintegration is performed in the body frame to avoid repropagation whenever system state is updated during optimization for computational efficiency.

The propagation is performed in body frame $\text{B}_k$ instead of the world frame as
\begin{equation}
    \label{eq:imu_propagation}
    \begin{aligned}
        \text{R}_\text{BW}^k \ {}_\text{W}\mathbf{p}^{k+1}        & = \text{R}_\text{BW}^k({}_\text{W}\mathbf{p}^{k} + {}_\text{W}\mathbf{v}^j \Delta{t_k} - \frac{1}{2}{}_\text{W}\mathbf{g}\Delta{t_k}^2) + \boldsymbol{\alpha}_k^{k+1} \\
        \text{R}_\text{BW}^k \ {}_\text{W}\mathbf{v}^{k+1}        & = \text{R}_\text{BW}^k({}_\text{W}\mathbf{v}^k - {}_\text{W}\mathbf{g}\Delta{t_k}) + \boldsymbol{\beta}_k^{k+1}                                                       \\
        \textbf{q}_\text{BW}^k \otimes \textbf{q}_\text{WB}^{k+1} & = \boldsymbol{\gamma}_k^{k+1}
    \end{aligned}
\end{equation}

where $\boldsymbol{\alpha}_k^{k+1}$, $\boldsymbol{\beta}_k^{k+1}$ and $\boldsymbol{\gamma}_k^{k+1}$ are the preintegration terms which only depend on the inertial measurements and biases. These preintegration terms along with their covariance ${}_\text{B}\mathbf{P}_k^{k+1}$ can be updated iteratively using discrete accelerometer and gyroscope measurements; see ~\cite{qin2017vins_mono, forster2016manifold} for details. The preintegrated terms $\boldsymbol{\alpha}_k^{k+1}$, $\boldsymbol{\beta}_k^{k+1}$ and $\boldsymbol{\gamma}_k^{k+1}$ are updated using the first-order approximation with respect to their biases if the bias estimates are relatively unchanged as proposed in ~\cite{forster2016manifold}. Otherwise, we redo the propagation using the updated bias estimates. The inertial error $\mathbf{e_i}^k$ is obtained from ~\eq{~\ref{eq:imu_propagation}} as

\begin{equation}
    \resizebox{.98\columnwidth}{!}{
        $
            \mathbf{e_i}^k =  \begin{bmatrix}\text{R}_\text{BW}^k ({}_\text{W}\mathbf{p}^{k+1} -  {}_\text{W}\mathbf{p}^{k} - {}_\text{W}\mathbf{v}^j \Delta{t_k} + \frac{1}{2}{}_\text{W}\mathbf{g}\Delta{t_k}^2) - \boldsymbol{\alpha}_k^{k+1} \\
                \text{R}_\text{BW}^k{}(_\text{W}\mathbf{v}^{k+1} -{}_\text{W}\mathbf{v}^k + {}_\text{W}\mathbf{g}\Delta{t_k}) - \boldsymbol{\beta}_k^{k+1}                                                          \\
                2[(q_\text{WB}^k)^{-1} \otimes q_\text{WB}^{k+1} \otimes (\boldsymbol{\gamma}_k^{k+1})^{-1}]_{xyz}                                                                                                  \\
                \mathbf{b}_g^{k+1} - \mathbf{b}_g^{k}                                                                                                                                                               \\
                \mathbf{b}_a^{k+1} - \mathbf{b}_a^{k}
            \end{bmatrix}
        $
    }.
\end{equation}

The preintegration term allows us to estimate the body orientation at time $j \in (k, k+1]$ during recursive updates which we will use to obtain the magnetometer error in the next section.

\subsection{Magnetometer Residuals}
\label{sec:magnetometer_residuals}

\begin{figure}
    \centering
    \vspace{0.1in}
    \includegraphics[width=0.85\columnwidth]{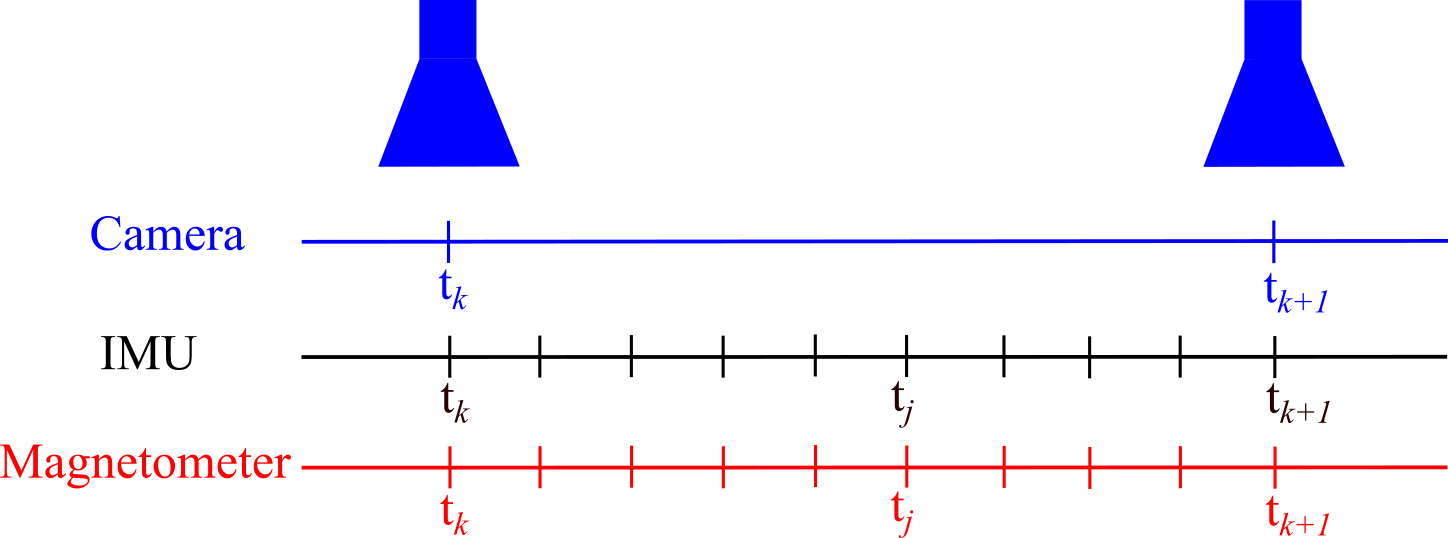}
    \caption{Illustration of visual, inertial and magnetometer measurement timing}\vspace{-0.3in}
    \label{fig:visual_inertial_magnetometer_timing}
\end{figure}

Magnetometer and IMU measurements obtained from 9-axis IMU used in attitude and heading reference systems (AHRS) are synchronized as shown in Fig.~\ref{fig:visual_inertial_magnetometer_timing}. The uncertainty of magnetometer measurements is modeled as Gaussian noise such that the magnetometer measurement is
\begin{equation}
    \label{eq:magnetometer_measurement}
    {}_\text{B}\tilde{\mathbf{m}} = \text{R}_\text{WB}^{\text{T}} \ {}_\text{W}\mathbf{m} + \boldsymbol{\eta}_m
\end{equation}
where ${}_\text{W}\mathbf{m}$ is the earth's magnetic field at the location in world coordinate frame which is aligned with earth's east, north and up (ENU) direction. The magnetometer additive white noise can be expressed as $\boldsymbol{\eta}_m \sim \pazocal{N}(\mathbf{0}, \sigma_m^2 \cdot \mathbf{I})$. We estimate the magnetometer white noise similar to IMU ~\cite{ieee_imu_standard} using the allan-variance plot.

However, the magnetometer measures the superposition of Earth's magnetic field and local magnetic field because of the presence of magnetic materials in the sensor's vicinity. Thus, the magnetometer is calibrated before experiments to estimate the soft and hard iron effects. The calibration procedure is explained in the next section. During VIO initialization, we align the orientation of the body frame in the ENU direction. When properly calibrated, the Earth's magnetic field points in the north-down direction. First, the upward-pointing gravity direction is obtained from the accelerometer measurements. The cross product of the magnetic field and the gravity direction provides the east direction. The remaining axis can be obtained from the cross product of the estimated axes. This is a common procedure used in AHRS systems such as the Madgwick filter~\cite{madwick_filter} and the complementary filter~\cite{complimentary_filter}. Formulating the magnetometer residual directly in terms of \eq{~\ref{eq:imu_measurements}} has one significant drawback that any error during the initial alignment is also included in the estimation process. Thus, we formulate the magnetometer residuals as relative orientation constraint between two consecutive frames.

For the magnetometer residual formulation, we assume that the magnetometer samples are calibrated following the procedure in ~\sect{~\ref{sec:magnetometer_calibration}}. Using the preintegrated orientation term allows us to use all magnetometer measurements between consecutive frames. Given the current consecutive frame state estimates $\mathbf{x}_k$ and $\mathbf{x}_{k+1}$, we define the magnetometer residual for measurement at time $t_j \in (t_k, t_{k+1}]$ as

\begin{equation}
    \label{eq:magnetometer_residual}
    \mathbf{e_m}^{j,k} = \text{R}_\text{BW}^{k} \ \text{R}_\text{WB}^{k+1} \ {}_\text{B}\tilde{\mathbf{m}}^{k+1} - {}\Delta\text{R}_k^j \ {}_\text{B}\tilde{\mathbf{m}}^{j}
\end{equation}

where $\Delta\text{R}_k^j$ is the propagated orientation between the time interval $[t_k, t_j]$ and transforms the magnetometer measurement from time $j$ to $k$. This propagated orientation is the rotation matrix representation of the intermediate preintegration result $\boldsymbol{\gamma}_k^{j}$. Rotation $ \text{R}_\text{BW}^{k} \ \text{R}_\text{WB}^{k+1} $ represents relative orientation during the time interval $[t_k, t_{k+1}]$ in the body frame $\text{B}_k$  and transforms the magnetometer measurement from time $t_{k+1}$ to $t_{k}$. By expressing the error term in the body frame, we can recursively calculate $\boldsymbol{\gamma}_k^{j}$ efficiently as $t_j < t_{k+1}$. Thus, the magnetometer error using preintegrated orientation allows us to use all the magnetometer measurements in the time interval $(k, k+1]$ to optimize the relative orientation between consecutive frames. The magnetometer residual does not depend on the position estimate; as such, it does not directly affect the state's position estimate. However, a better orientation estimate will certainly yield a better position accuracy in the long run.

The magnetometer measurement estimated at $t_k$ using quaternion preintegration depends on the gyroscope noise. However, the gyroscope noise is already considered while computing $\boldsymbol{\gamma}_k^{j}$ and is significantly smaller than the magnetometer noise. Also, we found that the propagated orientation variance was very small compared to the magnetometer variance. Thus, for simplicity and computational efficiency, we assume that the magnetometer residual weights depend only on the magnetometer sensor noise. Moreover, when the state connected to the magnetometer residuals is marginalized, the magnetometer residuals are added as the marginalization prior along with the inertial and visual residuals.

\subsection{Soft and Hard Iron Effect Calibration}
\label{sec:magnetometer_calibration}

The full magnetometer measurement model taking into account the magnetic disturbances and sensor imperfections can be modeled~\cite{mag_calib} as
\begin{equation}
    \label{eq:mag_full}
    {}_\text{B}\hat{\mathbf{m}} = \text{S} \text{R}_\text{WB}^{\text{T}} \ {}_\text{W}\mathbf{m} + \mathbf{h} + \boldsymbol{\eta}_m
\end{equation}

where S represents the soft iron matrix and $\mathbf{h}$ represents the hard iron effect. Hard iron effects arise because of the permanent magnetization of the material and also depend on the fixed sensor recording mechanism. In particular, we can think of the hard iron effect as a constant bias $\mathbf{h}$. Soft iron effects are due to the magnetization of ferromagnetic materials due to the local magnetic field and depend on the orientation with respect to the local magnetic field~\cite{magnetometer_inertial_calibration}. The soft iron effect can be represented as a 3$\times$3 symmetric matrix S. We do not consider other sources of magnetometer errors that could arise from non-orthogonality of the magnetometer axes or differences in sensitivity along the three magnetometer axes~\cite{mag_extra_calib}.

\begin{figure}[h]
    \centering
    \begin{tabular}{cc}
        \fbox{\includegraphics[height=0.17\textheight, trim={6.1in, 1in, 5in, 0.3in},clip]{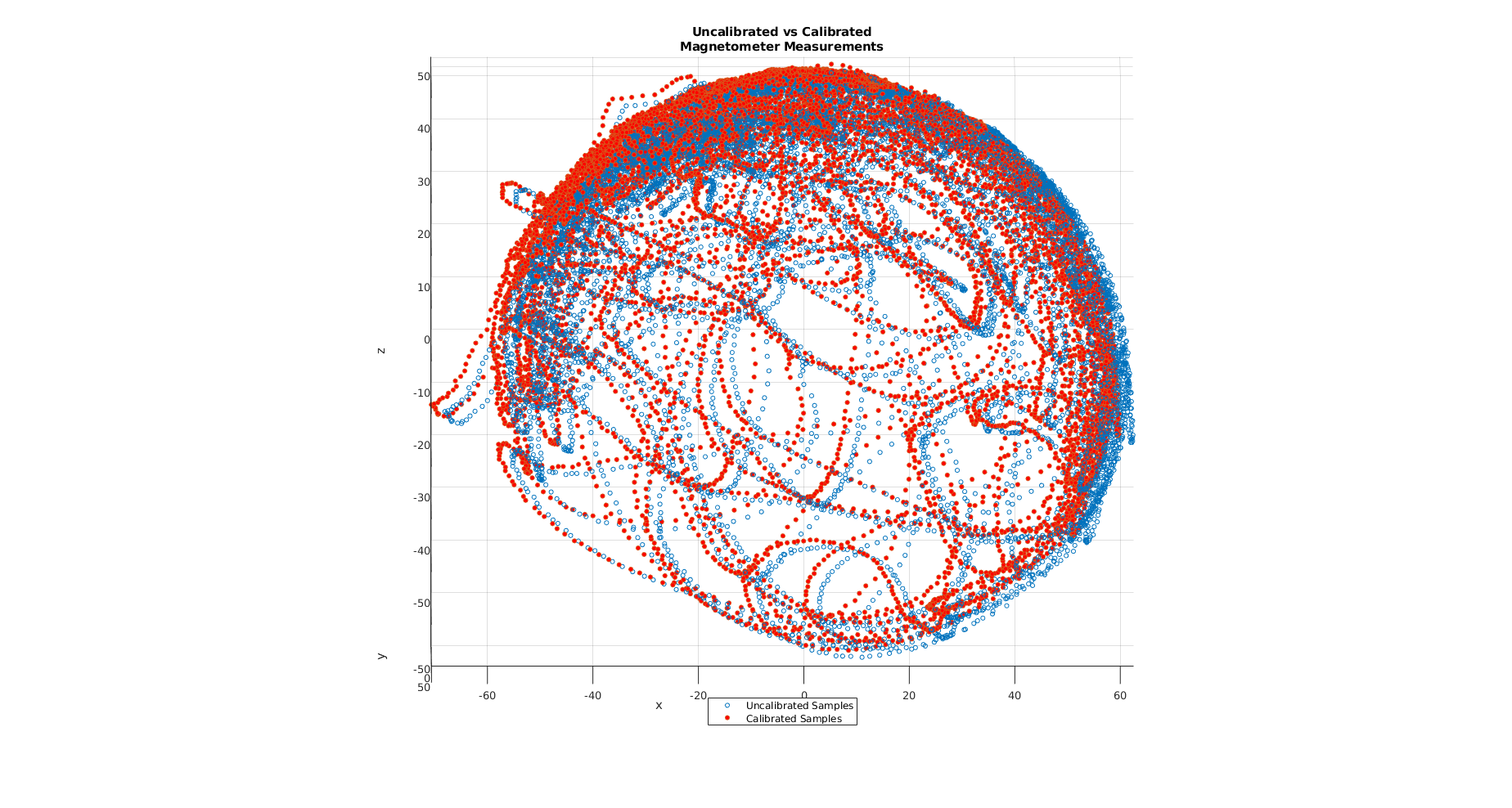}} &
        \fbox{\includegraphics[height=0.17\textheight, trim={6.1in, 1in, 5in, 0.5in},clip]{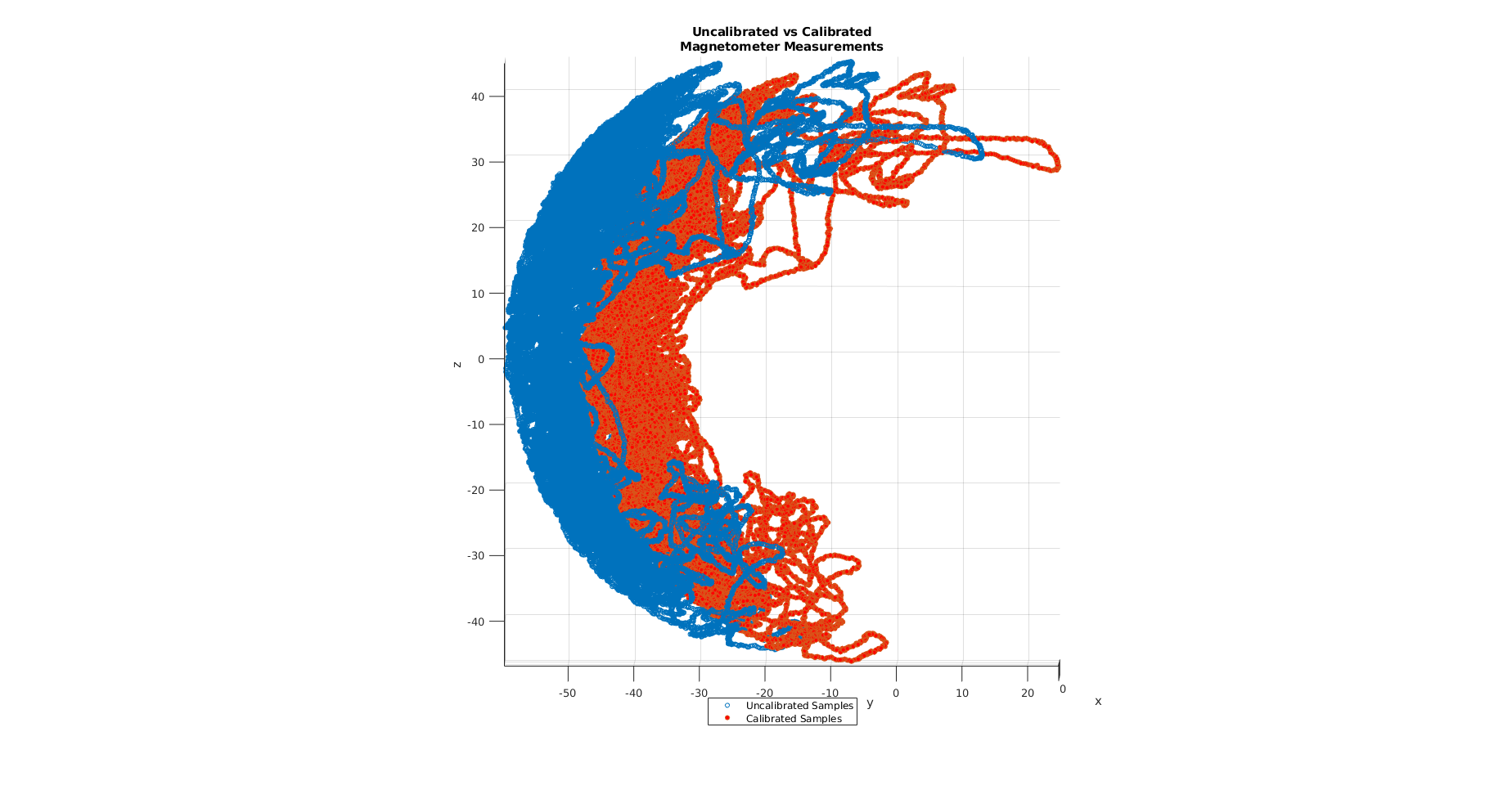}}
    \end{tabular}
    \caption{Magnetometer calibration ellipses fitted from explicit calibration sequence (left) and partial calibration (right).}\vspace{-0.2in}
    \label{fig:mag}
\end{figure}

Without the soft and hard iron offset, if we rotate a magnetometer sensor then the magnetic field vector falls on the surface of sphere with a radius equal to the magnitude of the earth's magnetic field. ~\eq{~\ref{eq:mag_full}} can be seen as a linear transformation that maps points on a sphere to an ellipsoid~\cite{mag_ellipsoid}. Thus, the magnetometer calibration problem can be interpreted as ellipsoid fitting of the points to the sphere surface. We solve the ellipsoid fit problem detailed in Kok \etal ~\cite{magnetometer_inertial_calibration} to obtain the estimate of $A = \text{S}^{-1}$ and the hard iron offset $\mathbf{h}$. For correct calibration, the magnetometer needs to be rotated along all three axes several times. Magnetometer measurements are corrected in the body frame as follows:
\begin{equation}
    \label{eq:mag_correction}
    {}_\text{B}\tilde{\mathbf{m}} = \text{A} ({}_\text{B}\hat{\mathbf{m}} - \mathbf{h})
\end{equation}
These corrected measurements are used to calculate magnetometer residuals in the sliding-window optimization as explained in ~\sect{~\ref{sec:magnetometer_residuals}}. In an indoor environment, these hard and soft iron effects continue to change as we move through the different areas. The underwater cave environment is ideally suited for magnetometer fusion as the likelihood of magnetic disturbance is minimal. As such, we performed the calibration at the beginning and assumed that the local magnetic field did not change significantly. During deployment, we performed calibration at the beginning and at the end of the dive, both calibrations resulted in similar values, validating the above assumption. \invis{To test the accuracy of the calibration, we check the radius of the sphere after ellipsoid fitting against the earth's magnetic field at that location.} On older datasets, we did not record an explicit magnetometer calibration sequence. In such a case, these are not enough magnetometer measurements in all directions; thus full mapping from ellipse to sphere is degenerate. Thus, we only calibrate the hard iron effect. Figure \ref{fig:mag} presents the ellipsoids fitted during complete calibration for the new sensor and for partial calibration of the old sensor.

\section{Experimental Results}
\label{sec:experiments}
We present evaluation results from two experiments in the Devil's Eye cave system, FL, USA and one experiment in the Dos Ojos Cenote, QR, Mexico. In all experiments the diver enters and exits the cave from the same location.

\subsection{Datasets}
\label{sec:datasets}
All the datasets are collected utilizing the sensor suite described in Rahman \etal~\cite{RahmanOceans2018}. The first dataset was collected using IDS UI-3251LE cameras in a stereo configuration and Microstrain 3DM-GX5-25 AHRS in the Devil's cave system, Florida. We call this dataset cave1\_short and images are captured at 15Hz. The cave1\_short dataset spans approximately 220m. The second dataset was collected with the same sensor suite in Dos Ojos Cenote, QR, Mexico. This trajectory is very long ($\sim$680m) with a duration of more than 50 minutes. COLMAP was unable to finish the complete trajectory due to computational and memory constraints. Thus, this data set is used only for qualitative evaluation.

The third dataset was collected using a similar sensor suite in monocular setup with Flir BFS-U3-16S7C-C camera and microstrain 3DM-GV7 AHRS. We call this dataset cave1\_long with total trajectory length of approximately 506m. Both datasets are very challenging in terms of low brightness, illumination changes due to artificial lighting, and low visibility.

\subsection{Results}
Due to the absence of GPS or motion capture systems underwater, COLMAP~\cite{colmap_sfm} is used to produce baseline trajectories. COLMAP is a Structure-from-Motion (SfM) framework that performs joint optimization of camera poses and map structure using global bundle adjustment. Since one of the datasets only contains monocular images, we consider the COLMAP trajectory up to scale for uniformity. Thus, the trajectories are aligned with the COLMAP baseline trajectories using \textit{sim3} alignment~\cite{umeyama}.

\begin{figure*}[h]
    \centering
    \vspace{0.1in}
    \begin{tabular}{lll}
        {\includegraphics[height=0.22\textheight]{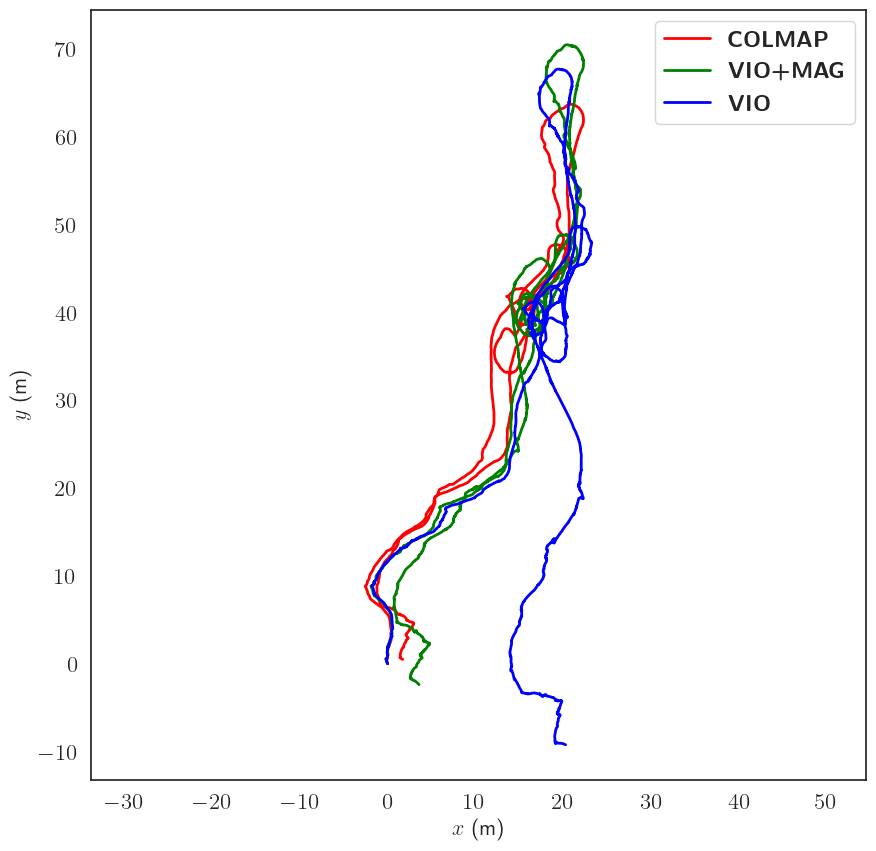}} &
        {\includegraphics[height=0.22\textheight]{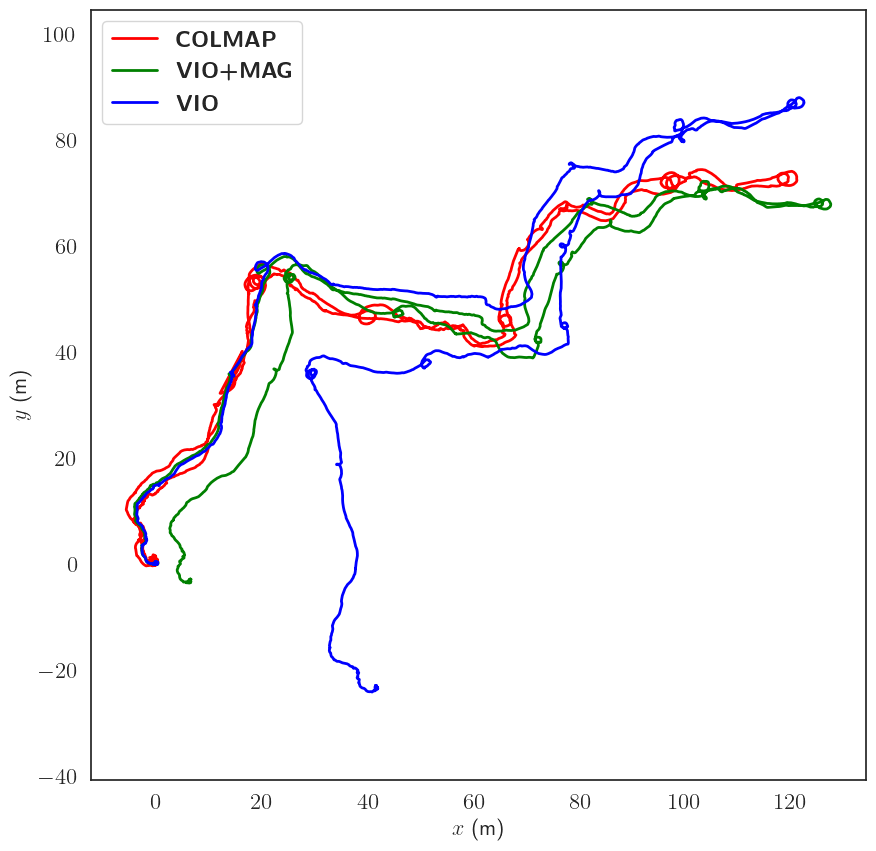}}  &
        {\includegraphics[height=0.22\textheight]{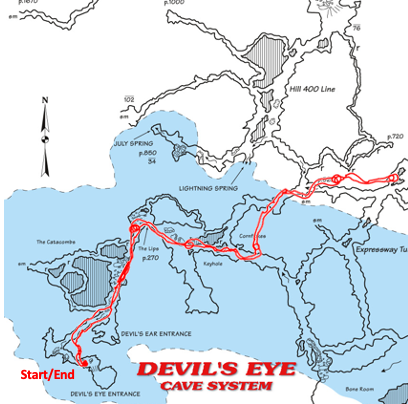}}
    \end{tabular}
    \caption{Trajectories from two deployments at the Devil's Eye cave system, FL, USA. The first two images show the trajectories, COLMAP (red) treated as ground truth, OKVIS VIO~\cite{okvis} (blue), and OKVIS with magnetometer (green) the proposed method. The first trajectory was carried out with the old sensor suite~\cite{RahmanOceans2018} and was a short ($\sim$220m) trajectory, the second was done with the new system and went further ($\sim$500m) trajectory. The final figure presents a section of the man\hyp made map with the ground truth trajectory superimposed (the blue background indicates the Santa Fe River over the cave). These trajectories are aligned at the origin (0,0) only to show endpoint error.}\vspace{-0.2in}
    \label{fig:Devil}
\end{figure*}
We compare the performance of our magnetometer formulation with the VIO formulation from OKVIS~\cite{okvis} in terms of absolute trajectory error (ATE) ~\cite{absolute_traj_error}. We include the results of the VIO-only case without fusing magnetometer measurements and compare them with the proposed method after \textit{sim(3)} alignment. Each method is run three times and the absolute trajectory error in terms of degree/meters is reported in Table \ref{tab:quantitative_results}. As seen in the table, there is a significant reduction in both translation and rotation RMSE. In particular in the cave1\_long dataset, both the rotation and translation error decreased significantly from $\sim$15\textdegree  to $\sim$6\textdegree and from $\sim$14m to $\sim$3m.

\begin{table}[h!]
    \centering
    {
        \caption{Three times run mean absolute trajectory error (ATE) for VIO with and without magnetometer compared with COLMAP baseline after \textit{sim(3)} alignment in terms of degree/meters.}
        \label{tab:quantitative_results}
        \begin{tabular}{@{}lccc}
            dataset      & length & VIO                     & VIO+MAG               \\
            \toprule
            cave1\_short & 220m   & 6.31\textdegree/6.45m   & 4.48\textdegree/4.98m \\
            cave1\_long  & 506m   & 14.65\textdegree/14.57m & 5.83\textdegree/3.37m \\
            \bottomrule
        \end{tabular}}

\end{table}

\begin{figure}
    \centering
    {\includegraphics[width=0.75\columnwidth]{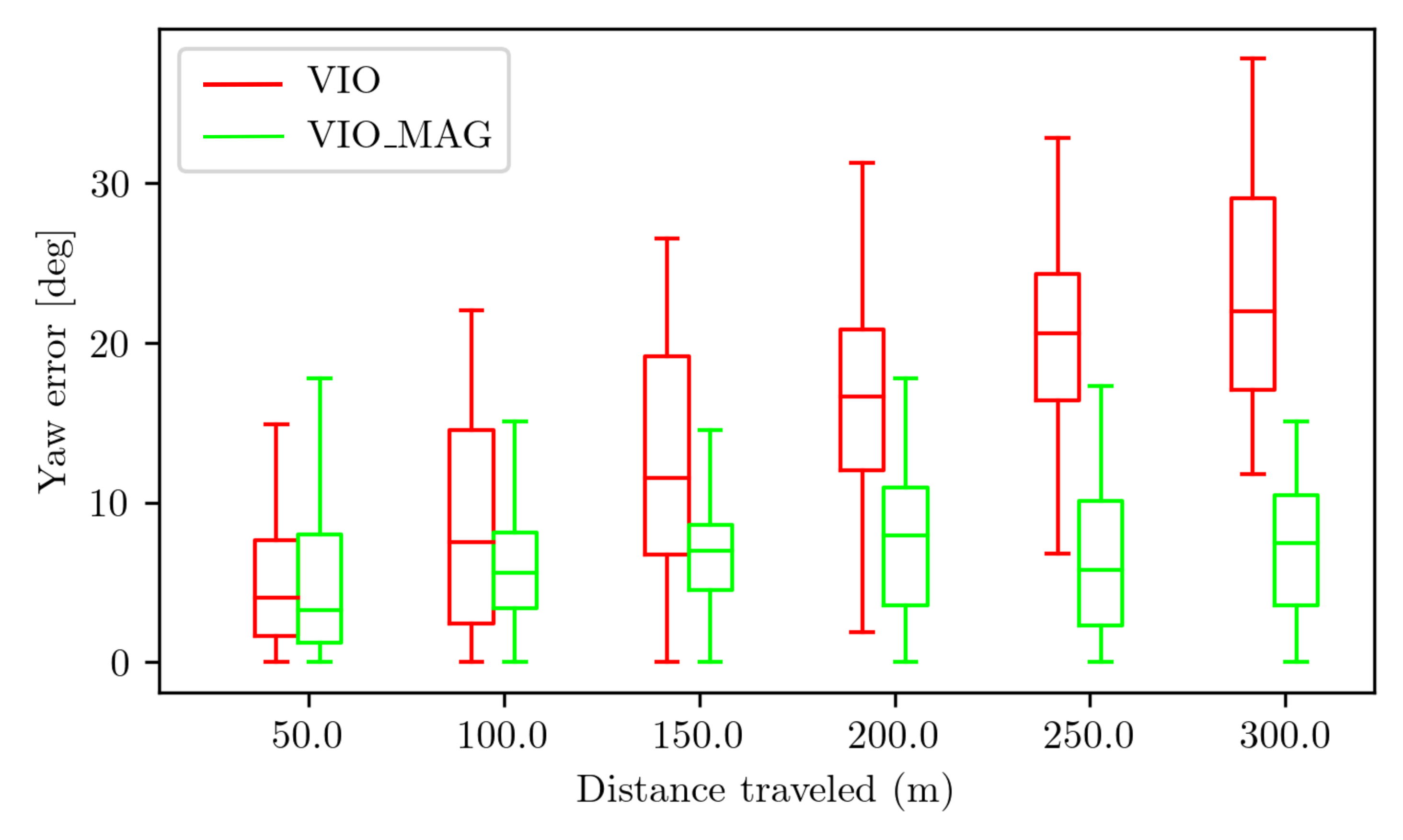}}
    \caption{Relative trajectory error (yaw) comparison between VIO and VIO+magnetometer.}\vspace{-0.25in}
    \label{fig:YawError}
\end{figure}

We also computed the relative trajectory error (RPE) that computes the relative error between states at different times with the focus on the yaw error ~\cite{Zhang18iros}. RPE shows the local drift over different sections of the trajectory and is unaffected by previously accumulated error. The relative yaw error calculated for cave1\_long trajectory after aligning the initial pose is shown in ~\fig{~\ref{fig:YawError}}. We can see that when using the magnetometer the yaw error remains constant over large distances, whereas the VIO accumulates error when distances keep increasing.

Two different deployments from the Devil's Eye cave system are presented in Fig. \ref{fig:Devil}. The first deployment, a short foray of 100 meters penetration inside the cave, can be seen in the left image, and the drift in the trajectory (blue) without magnetometer data fusion is clear. The visual inertial magnetometer fusion (green) follows much more closely the ground truth trajectory from COLMAP. More dramatic improvements are presented in the middle figure with trajectory of more than 500 meters. Compared to the globally optimized (COLMAP) trajectory there was reduction of positional error from 2.8\% to 0.6\%. The right image presents a partial map of the cave system with the COLMAP trajectory superimposed. The map is not up to scale, thus some passages are not scaled uniformly. Nonetheless, the trajectory follows the main passages of the cave system.

\begin{figure}
    \centering
    {\includegraphics[width=0.8\columnwidth]{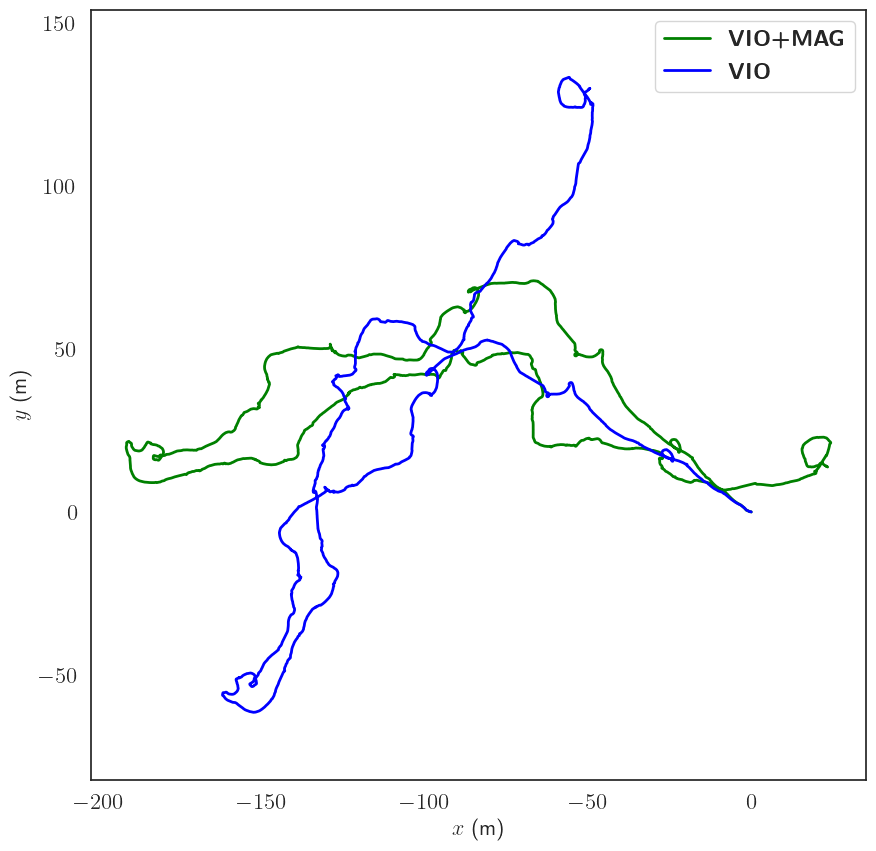}}
    \caption{Trajectories ($\sim$680m) from a deployment at Cenote Dos Ojos cave system, QR, Mexico. OKVIS VIO~\cite{okvis} (blue), and OKVIS with magnetometer (green) the proposed method. \invis{The experiment was performed with the old sensor suite~\cite{RahmanOceans2018} and it was a long ($\sim$680m) trajectory.} These trajectories are aligned at the origin (0,0) only to show endpoint error.}
    \label{fig:DosOjos}
    \vspace{-0.25in}
\end{figure}

Figure \ref{fig:DosOjos} presents the trajectory resulting from the deployment of the old sensor suite at the Cenote Dos Ojos. During the 50-minute deployment, the sensor covered approximately 680 meters inside a highly decorated cave. The large volume of data made the production of a COLMAP\hyp based ground truth trajectory infeasible due to computation and memory constraints. As such, this trajectory is used only as a qualitative comparison. As can be seen, the VIO diverged almost 100 meters. The trajectory obtained from magnetometer fusion follows the path out much more closely to the path taken on the way in. It is worth noting that the sensor's trajectory does not follow exactly the same path; however, the confined structure of the cave passage constrains the inbound and the outbound trajectories fairly close.

\section{Conclusion}
In this paper we presented a novel approach for augmenting VIO with magnetic field data. The experimental results demonstrated significant improvements in the accuracy of the resulting trajectories, significant reduction in the orientation error, and also the ability to orient the resulting trajectory with the magnetic north. During the estimation process, the magnetometer and the accelerometers provide absolute attitude information (with respect the magnetic north and the gravity vector).

Future work will explore the performance of the proposed system during shipwreck mapping, where the sensor traverses near metal structures. During initial calibration we expect to recover the true yaw of the sensor with respect to the magnetic north, enabling better positioning of the underwater structure in space. Furthermore, deployments of the proposed system on an autonomous underwater vehicle will require placing the magnetometer at some distance from the motors. A future experimental study will evaluate the magnetic noise levels for popular vehicles, such as the BlueRoV2~\cite{bluerov} or the Aqua2 AUV~\cite{DudekIROS2005}.

\newpage
\printbibliography

\end{document}